%% file: main.tex
\documentclass[11pt]{idea_style}
\newif\ifdraft
\draftfalse

\usepackage[utf8]{inputenc} 
\usepackage[T1]{fontenc}
\usepackage{url}
\usepackage{nicefrac}
\usepackage{multicol}
\usepackage{microtype}
\usepackage{geometry}
\usepackage{graphicx}%
\graphicspath{{fig/}}
\usepackage{multirow}%
\usepackage{amsmath}
\usepackage{amssymb}
\usepackage{amsfonts}
\usepackage{stmaryrd}
\usepackage{mathrsfs}%
\usepackage{xcolor}%
\usepackage{textcomp}%
\usepackage{manyfoot}%
\usepackage{booktabs}%
\usepackage{algorithm}%
\usepackage{tabularx}%
\usepackage{algorithmicx}%
\usepackage{algpseudocode}%
\usepackage{listings}%
\usepackage[inline]{enumitem}
\usepackage{xspace}
\usepackage{tikz}
\usepackage{makecell}
\usepackage{longtable}
\usepackage{changepage}
\usepackage{color}
\usepackage{colortbl}
\usepackage{pifont}
\usepackage{subcaption}
\usepackage{physics}
\usepackage{diagbox}
\usepackage{wrapfig}
\usepackage{siunitx}
\usepackage[misc]{ifsym} 
\usepackage[numbers]{natbib}
\usepackage{array}
\usepackage{xltabular}
\usepackage{ltablex}
\keepXColumns

\usepackage{tcolorbox}
\tcbuselibrary{skins, breakable}

\usepackage{hyperref}
\hypersetup{
  colorlinks,
  linkcolor=ideacolor,
  citecolor=ideacolor,
  urlcolor=ideacolor
}
\usepackage[nameinlink,noabbrev]{cleveref}

\usepackage{setspace}
\usepackage{changepage}
\usepackage{float}
\usepackage{tablefootnote}
\usepackage{threeparttable}
\usepackage{placeins}

\input{tex/math_commands.tex}

\input{tex/macros}

\ifdraft
    \providecommand\todo[1]{[\textcolor{red}{TODO: {#1}}]}
\else
    \providecommand\todo[1]{}
\fi

\newcommand{\system}{Mozi}

\newcommand{\mcp}{MCP}

\newtcolorbox{trajectorybox}[1]{
  enhanced,
  breakable,
  colback=blue!5!white,
  colframe=blue!75!black,
  coltitle=white,
  fonttitle=\bfseries,
  title={#1},
  boxrule=1pt,
  arc=3pt,
  left=10pt,
  right=10pt,
  top=10pt,
  bottom=10pt,
  fontupper=\small,
}

\setlist{nosep}

\newcolumntype{Y}{>{\raggedright\arraybackslash}X}

\author[]{He Cao$^\dagger$, Siyu Liu, Fan Zhang, Zijing Liu, Hao Li, Bin Feng, Shengyuan Bai, Leqing Chen, Kai Xie, Yu Li$^{(\textrm{\Letter})}$}

\affiliation{International Digital Economy Academy (IDEA)}
\ideadata[$^\dagger$]{Project Lead}

\ideadata[Corresponding author$^{(\textrm{\Letter})}$]{\url{liyu@idea.edu.cn}}
\ideadata[Main page]{\url{https://ai4s.idea.edu.cn/ai4s/mozi}}
\ideadata[Benchmark]{\url{https://huggingface.co/datasets/OpenMol/DD100}}
\title{{Mozi}\\[7pt]\Large Governed Autonomy for Drug Discovery LLM Agents}

\abstract{
\small{
\input{secs/1_abstract}
}
}

\begin{document}

\maketitle

\input{secs/2_introduction}

\input{secs/3_related_work}

\input{secs/4_problem_formulation}

\input{secs/5_system}

\input{secs/6_evaluation}

\input{secs/7_conclusion}


\bibliographystyle{unsrtnat}
\bibliography{main}

\appendix
\input{secs/8_append}

\end{document}

%% file: tex/math_commands.tex
\usepackage{amsmath,amsfonts,bm}

\def\eqref#1{equation~\ref{#1}}

\def\1{\bm{1}}

\DeclareMathAlphabet{\mathsfit}{\encodingdefault}{\sfdefault}{m}{sl}
\SetMathAlphabet{\mathsfit}{bold}{\encodingdefault}{\sfdefault}{bx}{n}



%% file: tex/macros.tex
\definecolor{myorange}{RGB}{220,128,0}
\definecolor{myblue}{RGB}{0,80,220}
\definecolor{myred}{RGB}{220,20,0}


%% file: secs/1_abstract.tex
Tool-augmented large language model (LLM) agents promise to unify scientific reasoning with computation, yet their deployment in high-stakes domains like drug discovery is bottlenecked by two critical barriers: unconstrained tool-use governance and poor long-horizon reliability. In dependency-heavy pharmaceutical pipelines, autonomous agents often drift into irreproducible trajectories, where early-stage hallucinations multiplicatively compound into downstream failures. To overcome this, we present \textbf{\system}, a dual-layer architecture that bridges the flexibility of generative AI with the deterministic rigor of computational biology. Layer A (Control Plane) establishes a governed supervisor--worker hierarchy that enforces role-based tool isolation, limits execution to constrained action spaces, and drives reflection-based replanning. Layer B (Workflow Plane) operationalizes canonical drug discovery stages---from Target Identification to Lead Optimization---as stateful, composable skill graphs. This layer integrates strict data contracts and strategic human-in-the-loop (HITL) checkpoints to safeguard scientific validity at high-uncertainty decision boundaries.

Operating on the design principle of ``free-form reasoning for safe tasks, structured execution for long-horizon pipelines,'' \system{} provides built-in robustness mechanisms and trace-level audibility to completely mitigate error accumulation. We evaluate \system{} on PharmaBench, a curated benchmark for biomedical agents, demonstrating superior orchestration accuracy over existing baselines. Furthermore, through end-to-end therapeutic case studies, we demonstrate \system{}'s ability to navigate massive chemical spaces, enforce stringent toxicity filters, and generate highly competitive in silico candidates, effectively transforming the LLM from a fragile conversationalist into a reliable, governed co-scientist.

%% file: secs/2_introduction.tex
\section{Introduction}

In the pharmaceutical industry, drug discovery is driven by the imperative to address unmet medical needs by producing safe, effective, and quality-controlled therapeutics. However, this process remains a protracted and capital-intensive endeavor, typically exceeding a decade and costing billions of dollars per approved drug. The canonical small-molecule drug discovery pipeline is a complex, multi-stage workflow: it begins with {Target Identification}, where disease-relevant biological mechanisms are validated; proceeds to {Hit Identification}, where vast chemical libraries are screened to find active compounds; advances to {Hit-to-Lead} and {Lead Optimization}, where molecular structures are iteratively refined for potency, selectivity, and pharmacokinetic properties; and culminates in rigorous preclinical and clinical evaluations \cite{seal2025ai}. Given the inherent complexity and high attrition rates of this pipeline, successful execution demands the orchestration of interdisciplinary expertise—spanning biology, chemistry, and pharmacology—and the seamless integration of heterogeneous data streams. Although high-throughput screening technologies and computational modeling have introduced efficiencies, the industry continues to grapple with fragmented workflows and the limitations of human cognition in processing large-scale multimodal data.

Recent advancements in artificial intelligence and machine learning have unlocked new possibilities for accelerating distinct nodes of this pipeline. A diverse array of specialized models has emerged, facilitating tasks such as structure prediction \cite{jumper2021af2, abramson2024af3, wohlwend2025boltz, passaro2025boltz2, baek2023rf2, corley2025rf3, ahdritz2024openfold, mirdita2022colabfold, bytedance2025protenix}, virtual screening \cite{alhossary2015fast, ding2023vina, tang2024vina, trott2010autodock, eberhardt2021autodock}, binding affinity estimation \cite{feng2025hierarchical, feng2024bioactivity, jia2026deep, ozturk2018deepdta, zhao2024pocketdta}, ADMET prediction~\cite{swanson2024admet, dong2018admetlab, xiong2021admetlab, fu2024admetlab}, and de novo molecular generation \cite{diffsbdd2024, peng2022pocket2mol, schneuing2025multi, guan20233d}. While these models operate with superhuman precision in isolated sub-tasks, they function as ``islands of intelligence,'' lacking the interoperability required for real-world pharmaceutical workflows. To bridge this gap, Large Language Model (LLM) agents have been proposed as cognitive orchestrators capable of connecting these disparate tools through reasoning and autonomous tool invocation. Early scientific multi-agent frameworks~\cite{gao2025pharmagents, li2025drugpilot, liu2024drugagent, inoue2025drugagent, pan2025frogent, jin2025stella, Jin2025BioLabEA, qu2025crispr} have demonstrated that mimicking human organizational roles (e.g., researcher, reviewer) can enhance success rates. However, a critical limitation persists: generic LLM agents are prone to probabilistic instability. In scientific rigor, unconstrained agents often suffer from tool-use hallucinations~\cite{Gridach2025AgenticAFA, schmidgall2025agent}, a lack of reproducibility, and an inability to adhere to strict Standard Operating Procedures (SOPs). Without a robust governance mechanism, the flexibility of agents becomes a liability, hindering their adoption in regulated enterprise environments where auditability and safety are paramount.

To address these challenges, we introduce \textbf{Mozi}, a novel agentic framework designed to reconcile the reasoning flexibility of LLMs with the rigorous, safety-critical demands of pharmaceutical R\&D. Unlike previous approaches that rely primarily on prompt-engineered role-playing, Mozi adopts a formalized \textbf{Dual-Layer Architecture} to achieve what we term \textit{governed autonomy}.
The architecture is structured into two symbiotic planes. First, at the strategic level, the \textit{Control Plane} (Layer A) acts as the governance engine. Instead of an unconstrained ReAct loop~\cite{yao2022react}, Layer A implements a hierarchical supervision protocol that translates high-level user intent into permissible actions. It enforces role-based access control to restrict tool usage based on agent clearance, manages dynamic error recovery through reflection, and maintains a strictly auditable trajectory of decision-making, ensuring that every algorithmic choice is traceable and compliant with organizational policies.
Second, underpinning the execution is the \textit{Workflow Plane} (Layer B), which materializes abstract scientific protocols into executable artifacts. We encapsulate the canonical drug discovery pipeline---from Target Identification to Lead Optimization---into composable, state-aware \textit{Skill Graphs}. Unlike independent tool calls, which often suffer from state-loss or I/O mismatch, these Skill Graphs function as directed acyclic graphs (DAGs) that enforce valid data flow and scientific logic (e.g., ensuring a protein structure is rigorously prepared before docking). By grounding the LLM's reasoning within these explicit graph-based constraints, Mozi mitigates the risks of hallucination and trajectory drift inherent in purely generation-based systems.

Experimental evaluations demonstrate that Mozi significantly enhances the reliability and interpretability of the automated drug discovery process. By effectively orchestrating a suite of domain-specific tools—including functional protein retrieval, structure-based docking, and ADMET prediction—Mozi achieves competitive performance in identifying high-quality candidates while maintaining full transparency. Crucially, our architecture integrates \textit{Human-in-the-Loop (HITL)} checkpoints at high-uncertainty decision boundaries, allowing expert scientists to intervene and guide the exploration process. This collaborative paradigm transforms the agent from a "black box" automation tool into a trustworthy co-scientist. In the following sections, we detail the architectural design of Mozi and present case studies to validate its efficacy in navigating complex, multi-objective optimization landscapes.

\begin{figure}[!htp]
\centering
\includegraphics[width=0.95\linewidth]{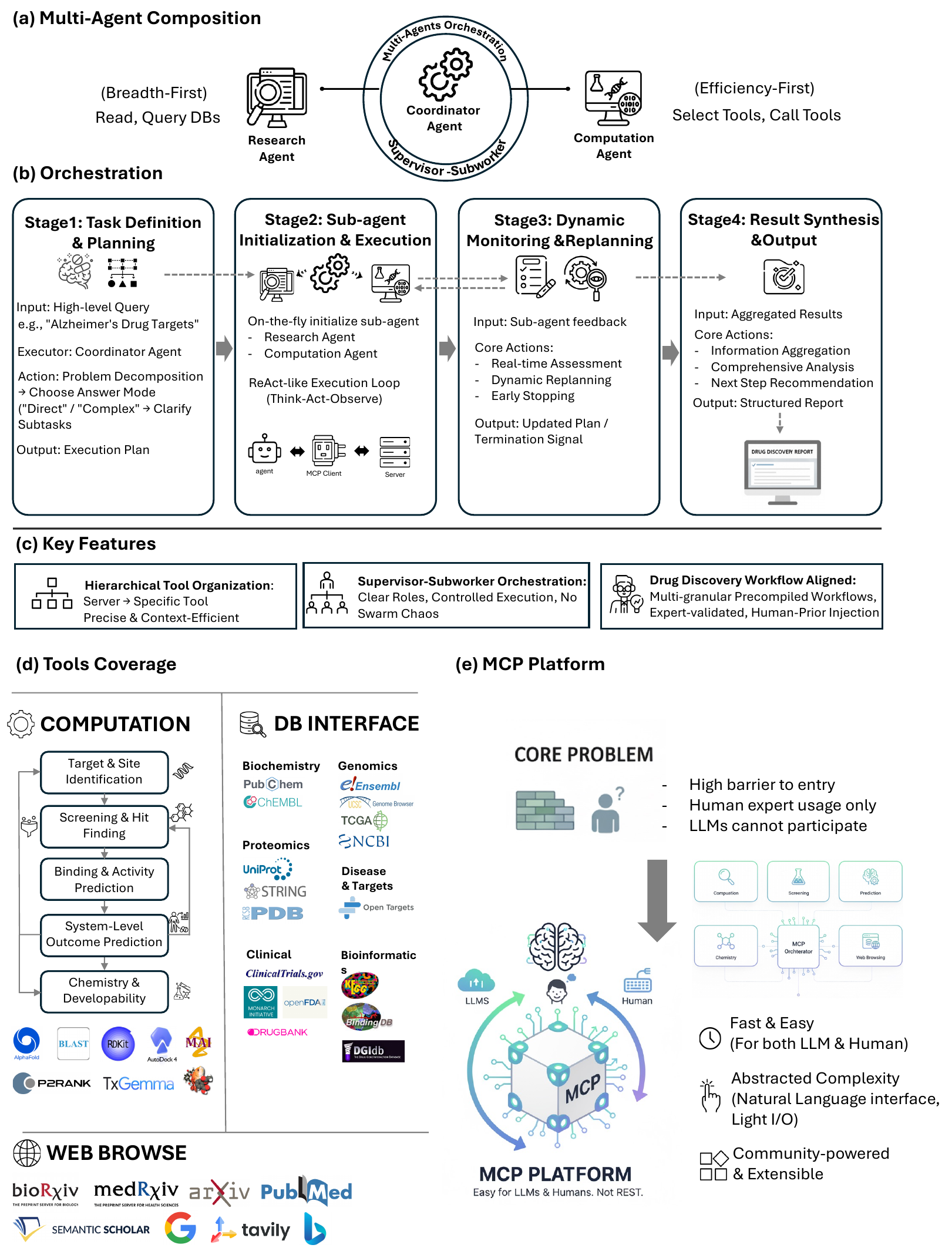}
\caption{System architecture of \system{} (a) Agent hierarchy featuring a central Coordinator managing specialized Research and Computation sub-workers. (b) The execution pipeline progresses through task definition, dynamic sub-agent instantiation, reflection-driven monitoring with replanning, and structured report synthesis. (c) Key design principles include hierarchical tool access, strict supervisor-subworker control, and alignment with multi-granular scientific workflows. (d) Coverage of integrated computational pipelines, biomedical databases, and real-time web retrieval tools is required for the canonical discovery lifecycle. (e) The MCP Platform serves as the foundational integration layer, abstracting complex computational biology tools into a standardized interface accessible to both autonomous agents and human experts.}
\label{fig:architecture}
\end{figure}

%% file: secs/3_related_work.tex
\section{Related Work}
\label{sec:related-work}

This section briefly positions \system{} among (i) tool-augmented agents and structured tool calling, (ii) hierarchical orchestration and reflection, (iii) tool governance and safety, (iv) workflow graphs/state machines for agents, and (v) AI systems for drug discovery pipelines.

\paragraph{From General Tool Learning to Scientific Action Spaces.}While general-purpose frameworks~\cite{qin2023toolllm, patil2024gorilla, yang2023gpt4tools, liu2024toolace} demonstrate that LLMs can learn to invoke APIs via fine-tuning or in-context learning, these methods often assume broadly semantic tools (e.g., calculators, calendars). However, scientific discovery demands interaction with high-stakes, domain-specific computational biology tools where parameter constraints are strict. In this domain, Biomni~\cite{huangbiomni2025} addresses this via a unified space of 150+ verified tools and code generation for complex logic. DrugPilot~\cite{li2025drugpilot} employs a parameterized memory pool and "Feedback and Focus" to enforce I/O constraints and reduce errors during multi-turn interactions. Weesep et al.~\cite{van2025exploring} note that while code-generating agents generally outperform direct tool-calling, they require prompt re-engineering when swapping models. We emphasize a constrained tool-use action space and bounded execution with fallbacks for verifiable stability.

\paragraph{Orchestration: From Chatbots to Scientific Pipelines.}
Multi-agent frameworks have standardized the "society of agents" metaphor for software engineering~\cite{hong2023metagpt, wu2023autogen, yang2024swe, wang2024openhands} and conversational tasks~\cite{li2023camel, schmidgall2025agent}. Yet, applying these generic architectures to drug discovery reveals gaps in long-horizon reliability and rigorous verification. Consequently, domain-specific hierarchies have emerged: PharmAgents~\cite{gao2025pharmagents} simulates a pharmaceutical company using a role-based hierarchy to optimize design workflows. DrugAgent~\cite{liu2024drugagent} employs a Planner-Instructor architecture to separate ideation from coding for ML automation. PharmaSwarm~\cite{song2025llm} uses a central Evaluator LLM to critique sub-agent proposals for hypothesis generation. Solovev et al.~\cite{solovev2024towards} show that multi-agent (Planner/Validator) architectures achieve 92\% success in complex neurodegenerative tasks where single agents fail. Our control plane operationalizes a bounded plan–execute–reflect loop with replanning, tailored for scientific tasks.

\paragraph{Tool Governance and Safety.}As agents gain tool access, governance (capability control, sandboxing, audit trails) becomes essential. Ünlü et al.~\cite{unlu2025auditable} propose an "auditable" platform recording reasoning provenance for molecular optimization to ensure regulatory transparency. DiscoVerse~\cite{zheng2025archives} uses multi-agent systems to securely mine confidential archives, preserving institutional memory while managing access. Seal et al.~\cite{seal2025ai} categorize safety implications, defining frameworks for "Perception, Action, and Memory" to ensure safe bounds. We adopt role-based isolation and treat side-effectful tools as explicitly permissioned, aligning capability with policy.

\paragraph{Workflow Graphs and State-Machine Control for Agents.}
Graph-based orchestration and state machines provide explicit control flow and recoverability, adding structure to LLM flexibility. PRIME~\cite{zhou2025prime} dynamically synthesizes DAGs from protein engineering tools to ground steps in verifiable execution. BioScientist Agent~\cite{zhang2025bioscientist} combines LLM orchestration with Knowledge Graph (KG) traversal via RL to elucidate mechanisms of action. Similarly, DrugAgent~\cite{inoue2025drugagent} integrates structured KG reasoning with text processing to improve drug-target interaction interpretability. We encode supervisor workflows and skills as graphs with explicit state interfaces for modular composition.

\paragraph{LLM Systems for Drug Discovery Pipelines.}
Drug discovery utilizes extensive computational pipelines, which LLMs now orchestrate. FROGENT~\cite{pan2025frogent} demonstrates an end-to-end agent executing tasks from target identification to retrosynthesis. In structure-based design, CIDD~\cite{gao2025pushing} uses LLMs to refine the "reasonability" of 3D molecules, bridging docking scores and chemical intuition. FRAGMENTA~\cite{suzuki2025fragmenta} utilizes agents to tune generative models via feedback loops autonomously. LIDDIA~\cite{averly2025liddia} and CLADD~\cite{lee2025rag} show how agents act as "digital twins" or use RAG to navigate chemical space without expensive fine-tuning. We focus on workflow-native design, emphasizing long-horizon reliability and HITL checkpoints at high-uncertainty points.

%% file: secs/4_problem_formulation.tex
\section{Problem Formulation}

We formulate the problem of LLM-assisted drug discovery not merely as a question-answering task, but as a \textit{governed, sequential decision-making process} over a heterogeneous tool environment. Unlike general-purpose assistants, a scientific agent must navigate a rigid landscape of dependencies where actions have high computational costs and side effects. In this section, we formalize the system dynamics, the governance constraints, and the reliability challenges inherent to long-horizon pharmaceutical workflows.

\subsection{Formalizing the Scientific Agent Workflow}
We define the drug discovery agent execution as a tuple $\mathcal{M} = \langle \mathcal{S}, \mathcal{T}, \mathcal{A}, \Pi \rangle$, operating over a discrete time horizon $t=0, \dots, T$.

\textbf{Hybrid State Space ($\mathcal{S}$).} The state of the system $s_t \in \mathcal{S}$ is hybrid, consisting of two distinct components: $s_t = (c_t, \mathcal{D}_t)$. Here, $c_t$ represents the unstructured interaction context (user intent, conversation history, and reasoning traces), while $\mathcal{D}_t$ represents the set of \textit{scientific artifacts} and external environment states. In the domain of drug discovery, $\mathcal{D}_t$ encompasses structured entities such as candidate target lists, verified Protein Data Bank (PDB) identifiers, localized binding pockets, and generated molecular structures (SMILES). Properly managing the synchronization between the reasoning context $c_t$ and the artifact state $\mathcal{D}_t$ is critical, as hallucinations in $c_t$ regarding $\mathcal{D}_t$ (e.g., referencing a non-existent PDB file) lead to execution failures.

\textbf{Tool Environment ($\mathcal{T}$).} The agent interacts with a set of domain-specific tools $\mathcal{T} = \mathcal{T}_{read} \cup \mathcal{T}_{compute} \cup \mathcal{T}_{write}$. These range from low-cost information retrieval (e.g., querying UniProt via $\mathcal{T}_{read}$) to high-cost computational simulations (e.g., molecular docking via $\mathcal{T}_{compute}$) and state-mutating operations (e.g., saving file artifacts via $\mathcal{T}_{write}$).

\textbf{Action Space ($\mathcal{A}$).} At each step, the agent selects an action $a_t \in \mathcal{A}$, which may involve generating a thought, invoking a tool $\tau \in \mathcal{T}$ with parameters $\theta$, or terminating the process.

\textbf{Execution Policy ($\Pi$)}. The agent acts according to a policy $\pi \in \Pi$, defined as $\pi(a_t|s_t)$, which governs the decision-making process by mapping the current hybrid state $s_t$ to a probability distribution over the action space $\mathcal{A}$.

\subsection{The Governance Challenge: Constrained Exploration}
A fundamental challenge in deploying LLMs to rigorous scientific domains is the mismatch between the unbounded generation capability of the model and the strict operational constraints of the laboratory environment. We define this as the Governance Problem.

Ideally, the agent should only select actions from a feasible region $\mathcal{A}_{\text{valid}} \subset \mathcal{A}$, defined by a set of hard constraints $\mathcal{C}$. However, standard LLM agents operate on the full space $\mathcal{A}$, often violating boundaries. We categorize these constraints into two types:
\begin{enumerate}[leftmargin=15pt]
    \item \textbf{Parameter Validity}: Tools in computational biology often require strict input formats (e.g., a specific chain ID in a PDB structure). Probabilistic agents frequently generate semantically plausible but syntactically invalid parameters, leading to execution crashes.
    \item \textbf{Role-Based Permission}: Not all agents should possess universal access. For instance, an agent responsible for literature search should not have permission to trigger computationally expensive docking simulations or overwrite existing experimental data.
\end{enumerate}
Therefore, the system must enforce a policy $\pi(a_t | s_t)$ that ensures $a_t \in \mathcal{A}_{\text{valid}}$ without severely hindering the reasoning flexibility of the model.

\subsection{Long-Horizon Reliability and State Dependency}
Real-world drug discovery workflows are characterized by deep causal dependencies between stages: Target Identification $\rightarrow$ Hit Identification $\rightarrow$ Hit-to-Lead $\rightarrow$ Lead Optimization. We model this as a directed graph of dependencies where the output of stage $i$ serves as the necessary input for stage $i+1$.

The primary technical difficulty is the \textit{propagation of error}. Let $\epsilon_t$ denote the uncertainty or error introduced at step $t$. In an unconstrained agentic loop, errors accumulate multiplicatively. A minor hallucination in the Target Identification stage (e.g., selecting a protein isoform irrelevant to the disease mechanism) renders all subsequent computations in Hit Identification—potentially costing hours of GPU time—scientifically invalid. 

Consequently, the system cannot treat the workflow as a single continuous context window. It requires a mechanism to discretize execution into verifiable stages, ensuring that the artifact state $\mathcal{D}_t$ satisfies specific quality gates before transitioning to the next pipeline phase.

\subsection{Objective: Auditable Trajectory Generation}
The overarching objective of the Mozi system is not merely to maximize a scalar reward function, but to generate a \textit{scientifically auditable trajectory} $\tau = (s_0, a_0, r_0, \dots, s_T)$. A successful trajectory must satisfy three criteria:
\begin{enumerate}[leftmargin=15pt]
    \item \textbf{Correctness}: The final molecular candidates and hypotheses must be grounded in valid evidence retrieved from $\mathcal{T}_{read}$ and validated by $\mathcal{T}_{compute}$.
    \item \textbf{Compliance}: Every action $a_t$ in the trajectory must adhere to the governance constraints $\mathcal{C}$ (budget, permission, and safety).
    \item \textbf{Reproducibility}: The sequence of decisions must be deterministic enough to allow human experts to audit the intermediate artifacts $\mathcal{D}_t$ and replay the logic.
\end{enumerate}
This formulation necessitates an architecture that separates high-level governance (Layer A) from standardized workflow execution (Layer B), as detailed in the following section.

%% file: secs/5_system.tex
\section{System Architecture}
To address the challenges of governance and long-horizon reliability defined above, we present the architecture of \system{}. The system is designed as a dual-layer topology that bridges the gap between the probabilistic reasoning of LLMs and the deterministic requirements of computational biology. As illustrated in the high-level design, \system{} consists of two symbiotic layers: the \textbf{Control Plane (Layer A)}, which acts as a hierarchical multi-agent system handling unstructured reasoning context ($c_t$); and the \textbf{Workflow Plane (Layer B)}, which manages structured scientific artifacts ($\mathcal{D}_t$) through stateful skill graphs. These layers are interconnected via the Model Context Protocol (MCP), a standardized bus that abstracts heterogeneous biological tools and databases into a unified service layer.

\subsection{Architectural Overview}
The separation of concerns between Layer A and Layer B is the foundational design principle of \system{}. Layer A is designed to preserve the flexibility of general-purpose agents for short-horizon interactions, knowledge retrieval, and dynamic planning. However, empirical observations in drug discovery reveal that for long-horizon, complex tasks, purely agentic planning often suffers from trajectory drift and accumulated errors, leading to unmanageable context lengths and local loops. Consequently, Layer B captures domain-specific Standard Operating Procedures as "Skill Graphs"—reusable, governed workflows that sacrifice local flexibility for rigorous, verifiable execution.

The execution flow begins with a \textit{Prompt-Based Intent Router} within Layer A. By applying rule-based logic via the LLM, the router classifies user intent into three categories: (1) \textit{Knowledge Retrieval}, handled directly via retrieval-augmented generation; (2) \textit{Single-Stage Tasks}, dispatched to a specific worker or skill module; and (3) \textit{End-to-End Workflows}, which trigger the Supervisor to orchestrate a sequence of skill graphs (e.g., Target Identification $\rightarrow$ Hit Identification).

\subsection{Layer A: The Governance and Orchestration Engine}
Layer A implements a Supervisor-Worker Hierarchical Agent System. Unlike flat multi-agent architectures, this hierarchy enforces strict command-and-control logic to maintain trajectory stability.

\paragraph{Supervisor Agent and Minimal Planning}
The Supervisor Agent serves as the central planner and constraint monitor. Upon receiving a complex request, the Supervisor generates a high-level plan consisting of minimal necessary steps. We reject open-ended exploration in favor of a bounded execution loop, as described in Algorithm~\ref{alg:supervisor}. The loop integrates a reflection mechanism—implemented as a self-correction prompt rather than a separate critic agent—to evaluate the completion status after each step. If a step fails or produces insufficient information, the Supervisor triggers a dynamic replanning routine rather than blindly continuing.

\begin{algorithm}[!htp]
  \caption{Supervisor loop (conceptual)}
  \label{alg:supervisor}
  \begin{algorithmic}[1]
    \Require user query $q$
    \State classify $q \rightarrow mode \in \{\text{direct},\text{simple},\text{complex}\}$
    \If{$mode = \text{direct}$} \State \Return completion \EndIf
    \If{$mode = \text{simple}$} \State select worker; run one episode; \Return result \EndIf
    \State plan $\leftarrow$ minimal\_steps$(q)$ with step budget $K$
    \For{$i = 1..|plan|$}
      \State $r_i \leftarrow$ run\_worker(plan[$i$], context=$\{r_{<i}\}$)
      \State decision $\leftarrow$ reflect($q$, completed=$\{r_{\le i}\}$, remaining=plan[$i{+}1..$])
      \If{decision = \texttt{SUFFICIENT\_INFO}} \State \textbf{break} \EndIf
      \If{decision = \texttt{REPLAN}} \State plan $\leftarrow$ update(plan, $i$, new\_steps) \EndIf
    \EndFor
    \State \Return synthesize($q$, completed=$\{r_{\le i}\}$)
  \end{algorithmic}
\end{algorithm}

\paragraph{Independent Workers and Role-Based Governance}
The Supervisor delegates tasks to specialized Workers (e.g., Research Worker, Computation Worker). Each Worker is instantiated as an independent agent with its own system prompt and isolated context window, ensuring that the token consumption remains localized. To solve the "Governance Problem," we implement governance not merely through prompting, but via Hard-Coded Tool Filtering. We define two isolation modes: \textit{Strict Mode}, used in production, which physically restricts the tool list available to a worker based on its role (preventing, for example, a Research Worker from accessing high-cost docking clusters); and \textit{Permissive Mode}, used for debugging, which exposes the full toolset. This prevents role overlap and ensures resource safety.

\subsection{Layer B: Artifact-Centric Skill Graphs}

While Layer A manages logic, Layer B manages the integrity of scientific artifacts. We encode long-horizon drug discovery workflows as Composable Stateful Skill Graphs. We define these skills as graphs rather than simple tool sequences because they require internal state management, parallel execution branches, and rigorous data schema enforcement. We implement Layer B using LangGraph, allowing us to encode long-horizon workflows as cyclic graphs with persistence and state management.

\begin{figure}[!ht]
\centering
\includegraphics[width=0.95\linewidth]{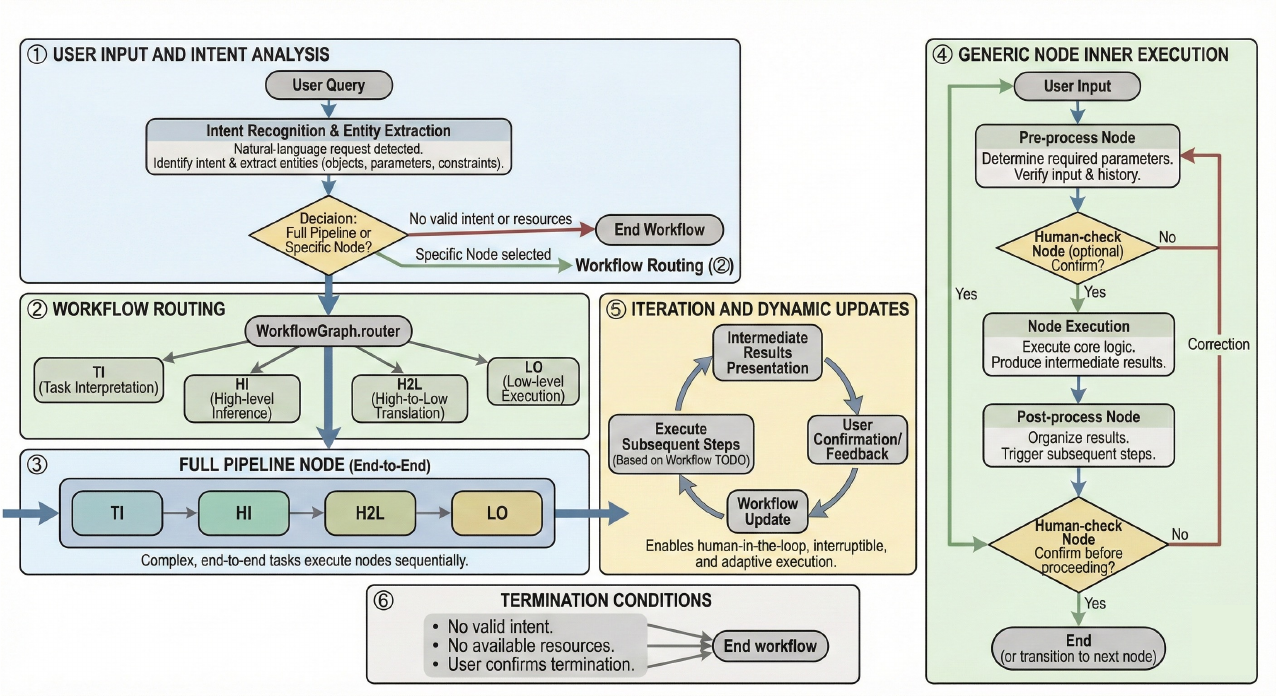}
\caption{Overview of the Layer B Workflow Plane. The framework parses user intent to route tasks to specific discovery stages or a full end-to-end pipeline (\textbf{1--3}). Within each step, a generic node execution mechanism (\textbf{4}) enforces rigorous input/output data contracts and embeds Human-in-the-Loop (HITL) validation gates. This stateful design supports dynamic iteration, expert intervention, and controlled termination (\textbf{5--6}) to mitigate long-horizon error propagation. See the details of each workflow in Appendix figures~\ref{fig:ti-workflow}, \ref{fig:hi-workflow}, \ref{fig:h2l-workflow}, and \ref{fig:lo-workflow}.}
\label{fig:skill-graph}
\end{figure}

\paragraph{Format Adaptation and State Contracts}
A persistent challenge in LLM-driven science is the instability of structured outputs. Layer B enforces explicit state contracts between nodes in the graph. We incorporate Format Adapters at the input and output of each node. These adapters programmatically validate and clean data—ensuring, for instance, that a PDB file adheres to standard formatting and is free of missing side chains—before passing it to sensitive computational tools. This prevents the "garbage in, garbage out" failure mode common in purely generative approaches.

\paragraph{Canonical Drug Discovery Skills}
We implement specialized graphs for each stage of the discovery pipeline, incorporating domain-specific strategies:
\begin{itemize}[leftmargin=15pt]
    \item \textbf{Target Identification (TI):} This graph automates the transition from a disease name to a valid structural input. It begins with entity normalization (mapping to MeSH/ICD) and aggregates multi-source intelligence from literature (e.g., PubMed) and clinical trials. An LLM agent scores and ranks potential targets based on evidence confidence. Upon selection, the graph automatically retrieves protein structures from the PDB and executes a Structure Preparation Protocol (e.g., using PDBFixer~\cite{eastman2017openmm}) to repair side chains and protonate residues, ensuring the artifact is simulation-ready.

    \item \textbf{Hit Identification (HI):} To maximize the discovery of active compounds, we implement a Parallel Dual-Stream Strategy. This design embodies a philosophy of complementary exploration: bridging the tension between chemical novelty and synthetic realizability.
    \begin{itemize}
        \item \textit{Path A (Generative):} Utilizes pocket-based generative models (e.g., DiffSBDD~\cite{diffsbdd2024}) to generate de novo molecules that fit the pocket geometry, followed by Tanimoto-based clustering to ensure diversity.
        \item \textit{Path B (Screening):} Executes High-Throughput Virtual Screening (HTVS) against commercial libraries using a deep learning-based interaction model (e.g., LigUnity~\cite{feng2025hierarchical}), which predicts affinity orders of magnitude faster than traditional docking while capturing non-linear interactions.
    \end{itemize}
    The results from both paths are fused, deduplicated, and re-ranked to produce a hit list that is both structurally diverse and experimentally actionable.

    \item \textbf{Hit-to-Lead (H2L):} This stage focuses on transforming validated hits into optimization-worthy leads through Chemical Space Expansion. The graph executes two parallel modification strategies: (1) R-Group Exploration to optimize side chains while maintaining the core scaffold, and (2) Scaffold Hopping to replace the core structure while preserving pharmacophores for intellectual property or physicochemical improvements. The expanded library undergoes a Two-Tier Filtration Process: first removing structural alerts (e.g., PAINS~\cite{baell2010new}, Brenk filters~\cite{brenk2008lessons}) and then applying quantitative ADMET prediction models (e.g., ADMET-AI~\cite{admetai2024}) to discard candidates with toxicity risks or poor bioavailability.

    \item \textbf{Lead Optimization (LO):} This stage employs a closed-loop Multi-Objective Optimization cycle. A Reinforcement Learning framework (e.g., REINVENT4~\cite{loefflerreinvent42024}) navigates the chemical space guided by a composite reward function that balances predicted affinity, drug-likeness (QED), and synthetic accessibility (SAS). Top candidates are periodically validated using high-fidelity physics-based methods (e.g., MM-GBSA). Finally, an LLM acts as a Virtual Medicinal Chemist, synthesizing the quantitative scores with qualitative insights from literature to generate the final ranked candidate report.
\end{itemize}

\paragraph{Human-in-the-Loop (HITL) Checkpoints}
To ensure scientific validity, we embed HITL gates as special nodes within the graphs. Unlike passive logging, these checkpoints pause execution at critical decision boundaries (e.g., before finalizing a candidate list for synthesis). The interface allows expert scientists not only to approve or reject results but also to perform {parameter correction} or trigger a {rollback} to a previous state. This feature aligns the autonomous execution with expert intuition, ensuring that the system remains a collaborative co-scientist.

\subsection{Data Fabric and Tool Federation}
The system's coherence relies on a robust data fabric that synchronizes the reasoning context with the artifact state.

\paragraph{Hybrid State Management}
The system maintains a hybrid memory structure. The Context State ($c_t$) utilizes rolling summaries of tool outputs to keep the LLM context window manageable. Simultaneously, the Artifact State ($\mathcal{D}_t$) tracks the actual files and data objects (SMILES lists, SDF files, docking grids) stored in the filesystem. Crucially, the system implements Provenance Tracking, recording the lineage of every artifact—capturing which agent, tool, and parameters generated it. This supports the reproducibility requirement of scientific experiments.

\paragraph{Tool Federation via MCP}
To manage the heterogeneity of biomedical tools—ranging from local Python scripts and Docker containers to remote cloud APIs—we utilize MCP. MCP provides a standardized abstraction layer that unifies discovery, invocation, and error handling. This allows Layer A to reason about tools as abstract functions while Layer B handles the complexities of execution, enabling the modular integration of new computational methods without disrupting the high-level orchestration logic.

%% file: secs/6_evaluation.tex
\section{Evaluation}
\label{sec:eval}

\subsection{PharmaBench: A Systems Benchmark for Drug Discovery Agents}
To rigorously assess the capabilities of \system{}, we propose \textbf{PharmaBench}, a curated benchmark of 88 tasks covering the full spectrum of drug discovery—from Target Identification to Preclinical Research. PharmaBench moves beyond simple retrieval or single-step prediction by incorporating scenarios that demand coherent scientific reasoning. The benchmark comprises three main sources: filtered tasks from the Therapeutics Data Commons (TDC)~\cite{huang2021therapeutics}, a dedicated subset of the Human-Last Exam (HLE)~\cite{phan2025humanity} focused on drug discovery, and auxiliary tasks from established databases or benchmarks~\cite{shen2024fine, cannon2024dgidb, buniello2025open}.

The quantitative foundation of PharmaBench consists of 55 tasks derived from the TDC. We selected these tasks to span both classification and regression modalities, covering critical subproblems such as drug--target interaction (DTI) prediction, ADMET profiling, toxicity assessment, and drug response quantification. This subset primarily evaluates the "Worker" layer of our architecture, specifically testing the system's ability to select appropriate computational tools (e.g., property predictors), parse structured biological data, and handle numerical regression with low error margins. We report Symmetric Mean Absolute Percentage Error (SMAPE) for regression tasks to account for the wide dynamic range of biological measurements, while standard Accuracy and F1 scores are used for classification and multiple-choice questions (MCQ). The inclusion of these tasks ensures that the agent provides robust, quantitatively valid inputs for downstream reasoning.

Complementing the quantitative tasks, we incorporated 28 HLE tasks (text-only) designed to stress-test the "Supervisor" layer and the high-level reasoning capabilities of the system. These tasks are not simple factoid retrievals; they are expert-written scenarios that simulate the complexity of laboratory decision-making and experimental troubleshooting. To ensure comprehensive coverage of the drug discovery lifecycle, we rigorously mapped each HLE task to a specific stage based on the biological entities and experimental contexts involved. Tasks involving gene-disease association logic, such as identifying virulence factors in pathogen genomes or selecting receptors for T-cell engineering, were categorized under Target Identification and Validation. Scenarios requiring the interpretation of enzyme kinetics, oligomeric state determination of coiled-coil sequences, or analysis of high-throughput screening data were mapped to Hit Identification and Lead Discovery. Furthermore, tasks involving the interpretation of in vivo mouse studies (e.g., ear swelling in antibody-drug conjugate trials) or analyzing pharmacokinetic profiles were classified under Preclinical Research. This manual taxonomy allows us to attribute performance gains specifically to improvements in stage-aware orchestration rather than generic reasoning improvements.

The final component includes 5 auxiliary tasks sourced from external benchmarks and databases to cover niche requirements such as protein function annotation and specific drug--gene interaction checks. Across all 88 tasks, the evaluation protocol is strict: we employ Exact Match metrics for string and numeric outputs where applicable, alongside manual verification for complex reasoning traces in the HLE subset. By combining the high-throughput, tool-heavy nature of the TDC subset with the reasoning-heavy, high-context nature of the HLE subset, PharmaBench provides a holistic view of an agent's readiness for deployment in a governed, scientific environment.

\subsection{Results}
We report results for \system{} under two representative model backends (labeled by their parameter scale in our logs) and compare against an existing biomedical agent baseline (Biomni\citep{huangbiomni2025}). We also report a small subset comparison of HLE, including several public systems (as recorded in our evaluation logs).

\paragraph{Results on \textsc{PharmaBench}}
Table~\ref{tab:bench} summarizes aggregate results on \textsc{PharmaBench}. Notably, \system{} improves overall classification/MCQ accuracy over the baseline in these runs.

\begin{table}[!htp]
  \centering
  \small
  \begin{tabularx}{0.72\linewidth}{lrrr}
    \toprule
    System & MCQ acc.$\uparrow$ & Class. acc.$\uparrow$ & Reg. (SMAPE)$\downarrow$ \\
    \midrule
    Biomni (Qwen3-30B) & 1/26 & 23/54 & 1.433 (8 tasks) \\
    \system{} (Qwen3-30B) & 2/26 & 25/54 & 1.420 (8 tasks) \\
    \midrule
    Biomni (Qwen3-235B) & 4/26 & 20/54 & 1.599 (8 tasks) \\
    \system{} (Qwen3-235B) & 6/26 & 33/54 & 1.169 (8 tasks) \\
    \bottomrule
  \end{tabularx}
  \caption{Results on \textsc{PharmaBench} (88 tasks).}
  \label{tab:bench}
\end{table}

Our logs include additional exact-match and F1 breakdowns for subsets of regression-like tasks. We include the full per-task log table in the project artifacts and recommend reporting per-subtask metrics in future versions of the benchmark.

\paragraph{HLE Drug Discovery Subset (28 tasks)}
We also report exact-match accuracy on 28 HLE drug discovery questions (see Table~\ref{tab:hle}). This subset emphasizes scientific reasoning and experimental design knowledge and is intentionally challenging; we include it primarily as a stress test for long-horizon scientific competence and tool-augmented behavior.

\begin{table}[!htp]
  \centering
  \small
  \begin{tabular}{lrr}
    \toprule
    System & Exact-match acc. & Correct / 28 \\
    \midrule
    Gemini-2.5-Pro (w/ web-search) & 10.71\% & 3 \\
    Biomni (Qwen3-30B) & 10.71\% & 3 \\
    Biomni (Claude-4.5-Sonnet) &17.86\%  &5 \\
    STELLA (Qwen3-30B) & 14.29\% & 4 \\
    SciMaster & 10.71\% & 3 \\
    MiroThinker & 14.29\% & 4 \\
    \midrule
    \system{} (Qwen3-30B) & 14.29\% & 4 \\
    \system{} (Qwen3-235B) & 17.86\% & 5 \\
    \system{} (Deepseek-V3.2) & \textbf{21.42}\% & \textbf{6}\\
    \bottomrule
  \end{tabular}
  \caption{Exact-match accuracy on 28 HLE Drug Discovery tasks.}
  \label{tab:hle}
\end{table}

\subsection{Analysis and Discussion}
The experimental evaluation primarily assesses system usability under low-cost conditions using open-source models. Performance on the TDC subset demonstrates that correct tool invocation directly yields accurate answers. This validates the system's ability to understand task intent, execute appropriate routing strategies, select relevant tools, and pass the correct parameters. For the HLE subset, \system{} powered by open-source models outperforms general-purpose configurations like Gemini-2.5-Pro~\cite{comanici2025gemini} augmented with web search. The results also indicate a clear scaling trend where employing more capable base models directly increases the overall accuracy of the agent system.

A qualitative comparison with other frameworks reveals distinct architectural trade-offs. Biomni~\cite{huangbiomni2025} achieves high accuracy only when supported by powerful proprietary models, suggesting an over-reliance on intrinsic model knowledge rather than robust agent validation mechanisms. Biomni's programmatic thinking philosophy drives it to write custom code for complex biological problems. This approach is brittle, as the agent often attempts to construct rule-based analysis pipelines from scratch for tasks like computing miRNA and target amino acid interactions. STELLA~\cite{jin2025stella} demonstrates ambiguous agent boundaries when operating on open-source models. The functional overlap between its development agent and tool creation agent causes redundant operations, such as repeated web searches across different nodes. Furthermore, the decision boundary between utilizing existing tools and generating new ones remains unclear. This ambiguity introduces significant execution risks where small models fail to properly evaluate tool quality, causing the system to enter infinite loops of creating and testing new tools.

Other systems face similar structural limitations. SciMaster~\cite{zhang2025bohrium+} employs a rigid operational workflow that attempts web retrieval before falling back to direct generation via the Intern-S1 model~\cite{bai2025intern}. The retrieval module frequently fails to extract precise information, forcing the system to rely entirely on the intrinsic scientific knowledge and computational reasoning capacity of the fallback model. MiroThinker~\cite{team2025mirothinker} adopts a machine learning engineer paradigm where it routinely downloads domain data to train custom models for specific queries. While its agentic search and code-based execution loop generally function well, this paradigm makes the problem-solving process exceedingly long. Consequently, even simple data retrieval tasks incur massive token consumption and time overhead. Additionally, parameter capacity limits in the underlying models occasionally cause repeated minor failures during tool calling, leading to localized execution loops.

\paragraph{Limitation.} The evaluation process also exposes several intrinsic limitations within our own architecture. The computation agent faces extreme difficulty in tool selection when the available domain tool registry scales to a massive number of options. When a tool invocation fails, the agent is prone to repeating the same tool call, creating localized execution loops that halt progress. Finally, computation agents driven by open-source models frequently exhibit parameter hallucination. This manifests as misaligned argument names, the unauthorized addition of non-existent parameters, and a fundamental misunderstanding of required data types during tool invocation.

\paragraph{What These Results Do and Do Not Measure}
The benchmark mixes tool-heavy prediction tasks (where correct tool selection and execution matter) and knowledge-heavy reasoning tasks. As such, aggregate accuracy conflates (i) model scientific priors, (ii) tool interface robustness, and (iii) orchestration correctness. We therefore complement benchmark results with long-horizon case studies that explicitly showcase stage-wise artifact generation and governance behavior.

\subsection{Long-Horizon Case Studies}
\label{sec:cases}

To evaluate the reliability and scientific coherence of the \system{} architecture, we report three end-to-end executions of the full workflow traversing target identification, hit identification, hit-to-lead expansion, and lead optimization. These execution traces demonstrate how the system maintains context across distinct stages, handles computational failures gracefully, and integrates human expertise at high-uncertainty decision boundaries.

The selection of Crohn's disease, Parkinson's disease, and sepsis provides a comprehensive evaluation environment covering diverse physiological mechanisms and pharmacological constraints. Crohn's disease is a chronic inflammatory condition of the gastrointestinal tract driven by immune dysregulation, which requires the modulation of complex protein interfaces to restore intestinal barrier function. Parkinson's disease is a progressive neurodegenerative disorder characterized by the deterioration of dopaminergic pathways, presenting severe structural design challenges because candidate molecules must successfully penetrate the blood-brain barrier while strictly avoiding neurotoxic side effects. Sepsis represents a life-threatening systemic inflammatory response to infection, necessitating the rapid and precise targeting of acute immunomodulatory receptors to stabilize vascular function and prevent organ failure. 

To contextualize the execution metrics presented in Table~\ref{tab:cases}, it is important to note the hardware setup and the algorithmic approximations utilized. The case studies were executed on a localized infrastructure utilizing 4x NVIDIA A6000 48GB GPUs and 128 CPU cores. The exceptionally high throughput observed in the Parkinson’s disease screening (377,760 compounds in ~35 minutes) was achieved by deploying a highly parallelized DTI model LigUnity~\cite{feng2025hierarchical} as an ultra-fast prefilter. This surrogate model infers binding affinity orders of magnitude faster than physics-based molecular docking, allowing the system to rapidly cull the vast library before committing intensive computational resources to the final structural validations.

\begin{table}[!htp]
  \centering
  \small
  \begin{tabularx}{\linewidth}{lYlrlrrr}
    \toprule
    Disease & Top-3 targets & PDB & \# Pockets & Hit Strategy \& Pool & \# Reps. & Dock fails & Time \\
    \midrule
    Crohn's & NOD2, ITGA4, IL12B & 3V4V & 42 & De novo (49) & 20 & 3 & 45m 24s \\
    Parkinson's & LRRK2, SNCA, GBA1 & 8TXZ & Ref. Ligand & HTVS (377,760) & 10 & 0 & 34m 33s \\
    Sepsis & ADRB2, TLR4, PTGS2 & 7BZ2 & 19 & De novo (48) & 20 & 7 & 48m 51s \\
    \bottomrule
  \end{tabularx}
  \caption{Summary of execution metrics for the three long-horizon demonstrations.}
  \label{tab:cases}
\end{table}

\paragraph{Crohn's Disease}
The system initiated a full discovery pipeline based on a general therapeutic request for Crohn's disease. Target identification retrieved 25 candidates and prioritized ITGA4 due to its strong clinical validation, selecting the PDB structure 3V4V, which represents an integrin heterodimer complex. Hit identification utilized the P2Rank algorithm to isolate a binding pocket and deployed the DiffSBDD generative model to design 49 novel compounds. Following structural clustering, the molecular docking phase evaluated 18 representative hits and identified a top molecule with a binding energy of -9.0 kcal/mol. The hit-to-lead stage expanded eight selected structures through parallel R-group exploration and scaffold hopping. The workflow applied stringent pharmacokinetic and toxicity filters, including PAINS, hERG blockade, and AMES mutagenicity checks to produce five lead molecules with zero penalty scores. Finally, lead optimization applied a reinforcement learning agent to maximize the quantitative estimate of drug-likeness and synthetic accessibility. This terminal stage yielded a final candidate molecule featuring a docking score of -8.8, combined with a near-perfect synthesizability profile.

\paragraph{Parkinson's Disease}
For the Parkinson's disease query, the system prioritized the LRRK2 kinase target and retrieved the high-resolution 8TXZ structure. During the hit identification stage, the human-in-the-loop checkpoint allowed the expert to override algorithmic pocket prediction and specify a legacy reference ligand to define the binding coordinates. The system subsequently screened a massive database of 377,760 compounds and isolated top hits based on deep learning drug-target interaction scores up to 0.972. The hit-to-lead stage expanded the top 10 candidates but detected persistent hERG channel blockade and hepatotoxicity risks across the expanded library, resulting in non-zero penalty scores. To resolve these severe liabilities, the lead optimization stage executed a multi-parameter optimization loop targeting specific safety constraints. The system successfully navigated the chemical space to discover a novel scaffold that demonstrated highly competitive in silico safety and blood-brain barrier permeability profiles relative to the established clinical benchmark DNL-201~\cite{jennings2022preclinical}, while maintaining a predicted strong binding affinity of -8.924 kcal/mol. We note that these values represent computational proxy metrics, and the true efficacy and safety profiles remain subject to future empirical validation.

\paragraph{Sepsis}
The therapeutic workflow for sepsis prioritized the ADRB2 receptor and selected the 7BZ2 cryo-EM structure. Hit identification located 19 potential binding pockets, generated 48 de novo molecules, and identified top binders with affinities reaching -8.7 kcal/mol. The hit-to-lead expansion processed 13 molecules through the standard toxicity and property filters to successfully yield five zero-penalty leads. During the lead optimization phase, the reinforcement learning agent generated thousands of molecular variations to maximize drug-like properties. The execution logs reveal that several generated molecules caused computational failures during the final AutoDock Vina validation step. Rather than terminating the workflow, the stateful graph logic gracefully captured these exceptions, bypassed the failed nodes, and penalized the problematic molecules in the final ranking. This robust error containment allowed the successful top candidates, which exhibited binding scores of -8.4 kcal/mol and excellent physicochemical properties, to proceed without requiring a full pipeline restart.

\paragraph{Analysis and Discussion of Execution Traces}
The detailed execution traces across these three distinct therapeutic areas highlight several critical advantages of the governed dual-layer architecture. The system demonstrates significant dynamic adaptability, seamlessly switching between generative artificial intelligence strategies for Crohn's disease and traditional high-throughput virtual screening for Parkinson's disease based on human guidance. Furthermore, the stateful workflow design provides necessary robustness against the stochastic failures common in computational biology. As evidenced in the sepsis case, localized Biodocking crashes are contained within the graph logic and do not propagate to the supervisory control plane. The Parkinson's disease execution specifically underscores the capacity of the system for multi-objective correction, where the agent recognized severe toxicity penalties in an intermediate stage and autonomously navigated toward a safer chemical scaffold during the final optimization cycle. Finally, the human-in-the-loop checkpoints prove essential for steering the autonomous process. These interactive validation gates enable domain experts to refine binding coordinates and verify target selections, ensuring that the computational trajectory remains aligned with clinical tractability before the system consumes significant computational resources.

\subsection{Comparative Analysis of Automated Drug Design Pipelines}

\begin{figure}[!ht]
\centering
\includegraphics[width=0.95\linewidth]{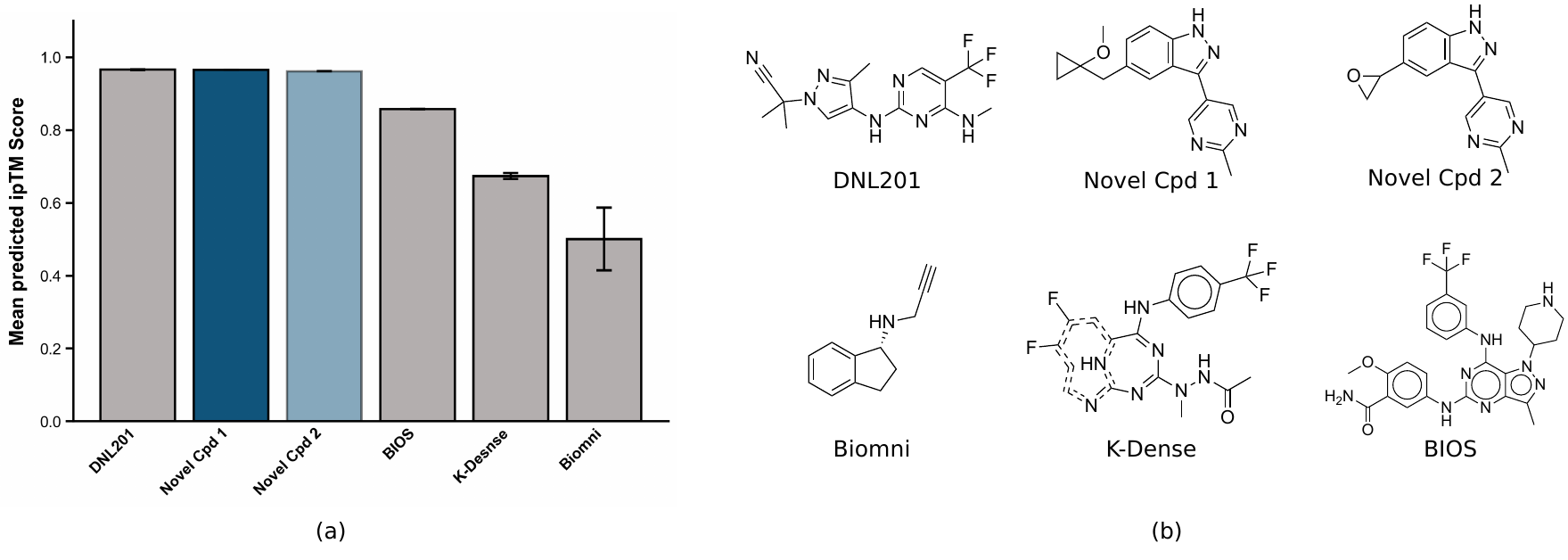}
\caption{Comparative analysis of LRRK2 inhibitors generated by different biomedical agent platforms. (a) Mean predicted AlphaFold3 Interface predicted TM-score (ipTM) for the Phase II clinical baseline DNL201, the top two novel compounds generated by our \system{}, and the top candidates from BIOS, K-Dense, and Biomni. (b) Two-dimensional chemical structures of the corresponding molecules.}
\label{fig:case-park}
\end{figure}

To further benchmark the end-to-end execution capabilities of \system{} against existing biomedical research platforms, we conduct a comparative analysis using Parkinson's disease as the target query. The evaluation focuses on designing inhibitors targeting LRRK2, using the 8TXZ protein structure as the standard input. We benchmark the generated molecules against DNL201~\cite{jennings2022preclinical}, a well-characterized LRRK2 inhibitor currently in Phase II clinical trials, which serves as the gold standard baseline. To ensure a fair and rigorous evaluation of binding affinity, we adopt the AlphaFold3~\cite{abramson2024af3} Interface predicted TM-score (ipTM) metric, following established methodologies in structural bioinformatics~\cite{hong2025good, gao2025helixdesign}. Figure \ref{fig:case-park} illustrates the predicted binding scores and the two-dimensional chemical structures of the top candidates generated by each platform.

The evaluation excludes the SciMaster~\cite{zhang2025bohrium+} platform because it provides only web search functionality and cannot execute end-to-end computational tasks. Similarly, the GaliLeo platform~\cite{GaliLeo} is omitted as it currently restricts operations to tool recommendation and single-tool execution without the capacity for automated task orchestration. The remaining baselines, Biomni~\cite{huangbiomni2025}, K-Dense~\cite{kdense}, and BIOS~\cite{bios}, operate as code-agent driven systems utilizing a deep research and code-based reasoning and acting paradigm. Their planning phases generally encompass target structure preparation, molecular library construction, ADMET-guided virtual screening, and final report generation. However, the Biomni platform failed to design novel molecules according to the prompt constraints. Instead, it relied on web search to retrieve Rasagiline, an existing clinical drug. Furthermore, attempts to execute molecular docking and ADMET prediction within Biomni failed due to backend environment instabilities. 

The K-Dense platform employs a multi-strategy library construction approach involving de novo generation and virtual screening. The de novo generation relies on fragment-based growth algorithms, encompassing substructure modification, side-chain decoration, and fragment combination, which yielded 580 unique compounds for docking. Nevertheless, the subsequent ADMET profiling within K-Dense is superficial, computing only basic physicochemical dimensions rather than comprehensive pharmacokinetic properties. Alternatively, the BIOS platform demonstrates rigorous preliminary research by identifying the mechanism of Type II LRRK2 inhibitors, establishing the GZD-824 scaffold as a validated template, and defining a comprehensive ADMET scoring system. This system integrates central nervous system multiparameter optimization, blood-brain barrier permeability (logBB), aqueous solubility (logS), P-glycoprotein efflux risk, Lipinski rule violations, and synthetic accessibility scores. Despite this theoretical rigor, the generation process remains completely opaque. Analysis of the output reveals a critical design flaw: the candidate molecules generated by the BIOS platform all exceed 500 Daltons in molecular weight, thereby violating fundamental physicochemical requirements for optimal central nervous system penetration. In contrast, the top two novel compounds generated by \system{} strictly adhere to the design constraints and achieve ipTM scores comparable to the DNL201 clinical baseline.

These comparative observations yield critical insights for long-horizon task automation in scientific domains. Systems driven purely by code-generating agents frequently waste computational resources by repeatedly reinventing standard operational procedures. While modern large language models demonstrate a strong intrinsic grasp of canonical drug discovery pipelines, successful end-to-end execution is heavily bottlenecked by the reliability and robustness of the underlying tool servers, particularly for high-concurrency and resource-intensive computations. This validates the architectural philosophy of \system{}, which mitigates execution drift by managing complex pipelines through highly controllable, stateful workflows coupled with robust backend tool federation.

%% file: secs/7_conclusion.tex
\section{Limitations and Risk Analysis}
\label{sec:limitations}

Despite its structured design and governance mechanisms, \system{} has several limitations that warrant careful consideration.

\begin{itemize}[leftmargin=10pt, noitemsep]
\item \textbf{Dependence on external tools and data sources.}
\system{} relies on heterogeneous external tools and databases accessed via standardized protocols. While protocolization improves interface consistency and provenance tracking, it cannot eliminate variability introduced by upstream data updates, API version drift, server availability, or changes in tool-side implementations. Such factors may affect reproducibility across runs or over time, particularly for long-horizon workflows that span multiple tool invocations.

\item \textbf{Residual stochasticity of LLM-based agents.}
Constrained action spaces, schemas, and execution budgets substantially reduce brittle or degenerate behaviors, but do not fully eliminate stochasticity in reasoning paths, tool selection, or result synthesis. As a consequence, different executions with identical inputs may still yield divergent intermediate artifacts or decisions, especially in early-stage exploratory tasks.

\item \textbf{Human-in-the-loop (HITL) trade-offs.}
HITL checkpoints improve reliability and safeguard high-impact decisions (e.g., entity confirmation or structure selection), but introduce additional user burden and reduce full automation. The optimal placement and frequency of HITL gates remain task-dependent and may vary with user expertise, computational cost, and acceptable risk levels.

\item \textbf{Scientific validity and safety.}
\system{} is designed as a decision-support system rather than an autonomous scientific authority. Generated hypotheses, rankings, or molecular candidates reflect model reasoning and tool outputs, and must be interpreted and validated by domain experts. Outputs should not be treated as clinical, therapeutic, or regulatory guidance.

\item \textbf{Benchmark and evaluation limitations.}
Existing mixed-format benchmarks conflate intrinsic model knowledge, tool interface robustness, and orchestration capability, making it difficult to isolate specific sources of success or failure. Although our evaluation highlights system-level behavior, more fine-grained benchmarks are needed to separately measure planning, tool selection, workflow execution, and scientific reasoning.

\item \textbf{Reliance on in silico surrogate models and lack of uncertainty quantification.} The evaluations in our case studies rely heavily on deep-learning proxy models (e.g., for DTI and ADMET predictions) and physics-based scoring functions (e.g., AutoDock Vina). These computational tools carry inherent predictive uncertainties, limited applicability domains, and lack calibration across unseen diverse targets. Consequently, any "benchmarking" against clinical compounds in this paper reflects algorithmic outputs rather than physiological realities. Future iterations of \system{} will need to integrate explicit Uncertainty Quantification (UQ) and cross-method consensus scoring to better inform the Human-in-the-Loop checkpoints before triggering expensive wet-lab syntheses.
\end{itemize}

\section{Conclusion and Future Work}
\label{sec:conclusion}

\subsection{Conclusion}
\system presents a workflow-native approach to LLM-assisted drug discovery by unifying a governed multi-agent control plane with a library of structured skill graphs for long-horizon scientific tasks. The proposed dual-layer architecture is designed to preserve the flexibility and generality of free-form LLM agents for short-horizon interactions, while introducing explicit structure, state, and governance for multi-stage discovery workflows. Through this separation, \system aims to improve reliability, auditability, and controllability without sacrificing broad scientific competence.

\subsection{Future Work}
Several directions remain open for future exploration. These include learning routing policies and automatically inducing reusable skills from execution traces, developing more formal and machine-checkable tool-governance mechanisms with stronger sandboxing guarantees, and integrating hybrid memory that combines explicit workflow state with long-term case-based retrieval. In addition, standardized evaluation suites that explicitly target governed, workflow-native scientific agents would enable more precise measurement of orchestration quality, robustness, and long-horizon reasoning performance.

%% file: secs/8_append.tex
\section{Tooling and MCP Integration}

\subsection{Database and Search Tooling}
\system{} integrates a large set of biomedical databases via \mcp. In our system instantiation, servers include (non-exhaustive): UniProt \citep{uniprot2019}, PubChem~\cite{pubchem2025}, ChEMBL~\cite{zdrazil2024chembl}, STRING~\cite{szklarczyk2023string}, Ensembl \citep{ensembl2025}, NCBI~\cite{sayers2025database}, PDB \citep{rcsbpdb2025}, Open Targets~\cite{buniello2025open}, DrugBank~\cite{knox2024drugbank}, KEGG~\cite{kanehisa2000kegg}, ClinicalTrials, and others. This provides unified access to protein knowledge, bioactivity data, structure metadata, and target--disease evidence.

\subsection{Computation Tooling}
We federate computational tools via \mcp{} as well, including (non-exhaustive):
\begin{itemize}[leftmargin=10pt, noitemsep]
  \item \textbf{Docking}: AutoDock Vina-based pose search and scoring.
  \item \textbf{Chemistry utilities}: RDKit and Open Babel \citep{openbabel2011} for SMILES operations and conversions.
  \item \textbf{Pocket localization / SBDD}: pocket finders and structure-based generation/optimization modules.
  \item \textbf{Property prediction}: ADMET predictors and task-specific ML models.
  \item \textbf{Generative chemistry}: hit-to-lead and lead-optimization pipelines (e.g., Reinvent-style workflows \citep{loefflerreinvent42024}).
\end{itemize}

\subsection{Governance via Server Provenance}
Tool isolation is enforced by mapping each tool to its source server and categorizing servers into \texttt{search}, \texttt{computation}, and \texttt{filesystem} types. In strict mode, each worker sees only its permitted tool category; in permissive mode exposes all tools for debugging.

\subsection{Tooling Description}

To support reliable, large-scale biomedical reasoning and workflow execution, our system integrates a comprehensive suite of external tools via MCP. These tools are explicitly divided into two disjoint categories: (i) database and search tools (\texttt{MCP-db\&search}) and (ii) computation tools (\texttt{MCP-computation}). This separation enables fine-grained governance, clear provenance tracking, and principled control over side effects during long-horizon workflows.

\paragraph{MCP-db\&search: Database and Retrieval Tools}

The \texttt{MCP-db\&search} layer provides read-only access to heterogeneous biomedical knowledge bases and is designed for structured information retrieval, entity grounding, and evidence collection. Tools in this category do not modify system state and are free of computational side effects, making them suitable for early-stage exploration and evidence aggregation.

Specifically, our system integrates major protein, chemical, genomic, and clinical databases, including UniProt (UniProtKB, UniRef, UniParc, proteomes, and gene-centric views), PubChem, ChEMBL, STRING, Ensembl, NCBI Gene and Genome resources, PDB, TCGA, UCSC Genome Browser, KEGG, Open Targets Platform, DrugBank, and ClinicalTrials. These tools expose standardized interfaces for tasks such as gene and protein lookup, compound and bioactivity retrieval, pathway and interaction querying, variant annotation, structural metadata access, disease--target association analysis, and clinical trial search.

All retrieval tools return structured outputs (e.g., JSON or tabular records) with explicit identifiers, database provenance, and version information. This design ensures that downstream reasoning steps can reliably reference entities, cross-link evidence across databases, and maintain auditability without relying on unstructured text scraping or heuristic parsing.

\paragraph{MCP-computation: Computational and Transformative Tools}

The \texttt{MCP-computation} layer federates tools that perform explicit computation, transformation, or model-based inference. Unlike database tools, these components may incur nontrivial computational cost and produce derived artifacts, and are therefore governed separately within the system.

Representative computation tools include molecular docking and scoring pipelines (e.g., AutoDock Vina-based workflows), cheminformatics utilities for molecular representation and conversion (RDKit and Open Babel), pocket detection and structure-based drug design modules, property and ADMET prediction models, and generative chemistry pipelines for hit-to-lead and lead optimization (e.g., Reinvent-style workflows). These tools operate on inputs that are typically grounded and validated via the \texttt{MCP-db\&search} layer, such as protein structures, binding sites, or curated compound sets.

To mitigate failure propagation and uncontrolled tool usage, computation tools are invoked under explicit supervision with bounded execution budgets, structured input schemas, and result validation. Outputs that exceed context limits are summarized according to the system’s long-output summarization policy, while raw artifacts are retained externally as part of the execution trace.

\paragraph{Design Rationale}

The explicit separation between \texttt{MCP-db\&search} and \texttt{MCP-computation} reflects a core design principle of our system: retrieval and reasoning should be decoupled from expensive or side-effectful computation. This distinction enables role-based tool isolation, simplifies auditing and debugging, and supports reproducible, stage-wise execution in long-horizon drug discovery workflows.

\subsection{Server Descriptions}
\label{app:server-descriptions}

{ 
\small 
\setlength{\tabcolsep}{6pt}
\renewcommand{\arraystretch}{1.3} 

\captionsetup{width=\textwidth, margin=0pt, labelfont=bf}
\begin{xltabular}{\textwidth}{
>{\hsize=0.5\hsize\raggedright\arraybackslash}X 
>{\hsize=1.0\hsize\raggedright\arraybackslash}X 
>{\hsize=1.5\hsize\raggedright\arraybackslash}X
}
    \caption{MCP server descriptions used in our system. Each server exposes a coherent toolset with a clear strategic role in long-horizon drug discovery workflows.} \label{tab:mcp-server-descriptions} \\
    
    \toprule
    \textbf{MCP server} & \textbf{Toolset overview} & \textbf{Core purpose \& key functionalities} \\
    \midrule
    \endfirsthead

    \caption[]{MCP server descriptions (Continued)} \\
    \toprule
    \textbf{MCP server} & \textbf{Toolset overview} & \textbf{Core purpose \& key functionalities} \\
    \midrule
    \endhead

    \midrule
    \multicolumn{3}{r}{\textit{Continued on next page}} \\
    \endfoot

    \bottomrule
    \endlastfoot

    \texttt{biorxiv} &
    Programmatic interface to query \textit{bioRxiv} / \textit{medRxiv} via \texttt{api.biorxiv.org}, enabling search, retrieval, and monitoring of preprints. &
    \textbf{Purpose:} automate literature surveillance for early-stage discovery (target validation, competitive intelligence). \newline
    \textbf{Functions:} (i) retrieve a specific preprint by DOI (title/authors/abstract/category/date); (ii) track publication status (map preprint DOI to published DOI, journal, date); (iii) time-based browsing (date range, recent days); (iv) category-filtered search with result limits. \\

    \texttt{chembl} &
    Comprehensive interface to \textit{ChEMBL} for bioactive molecules, targets, assays, and curated bioactivity measurements relevant to drug discovery. &
    \textbf{Purpose:} streamline target validation and hit/lead identification; enable scriptable access to SAR-relevant evidence. \newline
    \textbf{Functions:} bioactivity/assay retrieval (IC\textsubscript{50}, K\textsubscript{i}, etc.); compound/drug/biotherapeutic profiles (properties, indications, MoA, safety); target exploration (classes, components, binding sites); structure search (similarity/substructure, 2D depictions); ontology/classification (ATC/GO/protein families); supporting metadata (documents, organisms, tissues, cell lines, releases). \\

    \texttt{clinicaltrials} &
    Interface to \textit{ClinicalTrials.gov} API for trial search and detailed study retrieval via NCT identifiers. &
    \textbf{Purpose:} clinical landscape analysis and competitive intelligence (monitor studies, phases, recruiters, sponsors, interventions). \newline
    \textbf{Functions:} keyword/condition/location-based search; fetch detailed trial record by NCT ID (title, status, phase, sponsor, interventions, protocol description). \\

    \texttt{drugbank} &
    Interface to \textit{DrugBank} for drug/target knowledge, pharmacology, and drug--drug interactions. &
    \textbf{Purpose:} accelerate automated drug intelligence (MoA, indications, interactions) for hypothesis generation and safety checks. \newline
    \textbf{Functions:} drug search (name/indication/category); retrieve drug details by DrugBank ID (description, MoA, indication, identifiers); drug--drug interaction queries with effect descriptions. \\

    \texttt{ensembl} &
    Interface to \textit{Ensembl} REST API for multi-species genomic features, sequences, variants (VEP), and comparative genomics. &
    \textbf{Purpose:} target identification/validation using genomic context, variants, conservation, and disease associations. \newline
    \textbf{Functions:} gene/transcript/protein retrieval; sequence fetch (DNA/cDNA/protein, loci); variant effect prediction (VEP; SIFT/PolyPhen); homology/trees/gene-family evolution; phenotype--genotype links; mapping/xrefs/assemblies/species; LD and population genetics utilities. \\

    \texttt{kegg} &
    Interface to \textit{KEGG} to search, retrieve, and cross-reference genes, pathways, diseases, and compounds via API. &
    \textbf{Purpose:} provide pathway-level biological context for MoA, on/off-target reasoning, and disease mechanism grounding. \newline
    \textbf{Functions:} database metadata; keyword search/listing; entry retrieval (e.g., pathways, compounds; KGML); cross-references and identifier conversion (e.g., KEGG $\leftrightarrow$ NCBI). \\

    \texttt{ncbi} &
    Programmatic access to \textit{NCBI} resources (genes, genomes, viruses, taxonomy) for foundational biological metadata and annotations. &
    \textbf{Purpose:} support target validation by retrieving gene function/GO, orthologs, genomic context, and organism taxonomy. \newline
    \textbf{Functions:} gene lookup by multiple identifiers; genome/virus dataset reports and sequence/annotation access; taxonomy queries and subtree retrieval; BioSample and structured report generation. \\

    \texttt{opentarget} &
    Interface to \textit{Open Targets Platform} for integrated evidence linking targets, diseases, and drugs with quantitative scores. &
    \textbf{Purpose:} target prioritization and de-risking via evidence-based target--disease associations. \newline
    \textbf{Functions:} entity search/retrieval (targets/diseases/drugs); target$\leftrightarrow$disease association mapping with scores and evidence summaries. \\

    \texttt{pdb} &
    Suite for interacting with \textit{Protein Data Bank (PDB)}: structure download, component/assembly/interface queries, and external cross-references. &
    \textbf{Purpose:} automate structural data acquisition for structure-based discovery and mechanistic analysis. \newline
    \textbf{Functions:} download PDB files; entry metadata; polymer/ligand entities and instances; biological assemblies and polymer interfaces; annotations via PubMed/UniProt/DrugBank links; grouping/clustering (e.g., sequence similarity). \\

    \texttt{pubchem} &
    Interface to \textit{PubChem} for compounds/substances/bioassays, supporting rich identifier-based querying (CID/SID/AID) and substructure search. &
    \textbf{Purpose:} automate chemical intelligence and early ADMET/activity evidence gathering at scale. \newline
    \textbf{Functions:} name/SMILES/formula/CID/SID searches; property and synonym retrieval; assay summaries and AID details; links to genes/proteins; substructure-based queries. \\

    \texttt{pubmed} &
    Interface to \textit{PubMed} for literature search, summaries/abstracts, and related-article discovery via PMIDs. &
    \textbf{Purpose:} evidence collection for target validation and decision-making across discovery stages. \newline
    \textbf{Functions:} keyword and author-based search; fetch article summary and abstract by PMID; retrieve related publications for expansion. \\

    \texttt{rdkit} &
    Computational chemistry toolset powered by \textit{RDKit} for molecular descriptors, fingerprints, similarity, and structure manipulations from SMILES. &
    \textbf{Purpose:} automated small-molecule analysis for screening and lead optimization (drug-likeness, similarity, standardization). \newline
    \textbf{Functions:} property/descriptors (MW, LogP, TPSA, rings, HBD/HBA, rotatable bonds, QED); conversions (SMILES/SDF/InChI); stereo/aromaticity handling; fingerprinting (Morgan/MACCS/etc.) and similarity; substructure/ring/fragment analysis and clustering. \\

    \texttt{string} &
    Interface to \textit{STRING} for protein--protein interaction (PPI) networks, identifier mapping, and enrichment/annotation. &
    \textbf{Purpose:} mechanism-of-action and systems-level target validation via interaction networks and functional enrichment. \newline
    \textbf{Functions:} ID mapping; build/expand PPIs with confidence filters and interaction types; enrichment (KEGG/GO/Pfam/InterPro); annotation retrieval; homology/best-hit analysis across species. \\

    \texttt{tcga} &
    Focused toolset for gene expression analysis across \textit{TCGA} cohorts (via Firebrowse), identifying aberrant expression patterns. &
    \textbf{Purpose:} oncology target prioritization by rapidly profiling expression across cancer types. \newline
    \textbf{Functions:} retrieve expression across cohorts; identify significantly high/low expression cancers using z-score thresholds. \\

    \texttt{tooluniverse} &
    Integration hub that dynamically loads multiple biomedical toolsets into a unified MCP server and provides higher-level orchestration meta-tools. &
    \textbf{Purpose:} reduce data fragmentation by chaining multi-database calls for disease--gene--drug--phenotype intelligence. \newline
    \textbf{Functions:} multi-database integration; search and ID mapping; evidence retrieval; association analysis (target--disease--drug); meta-tools that compose workflows into single calls; similarity/relational discovery via literature-trained models. \\

    \texttt{ucsc} &
    Interface to \textit{UCSC Genome Browser} API for genomes/tracks discovery, sequence retrieval, and region-based annotation extraction. &
    \textbf{Purpose:} provide genomic neighborhood context for targets and disease loci in automated genomics workflows. \newline
    \textbf{Functions:} list assemblies/tracks/hubs/chromosomes; fetch sequences for loci (incl. reverse complement); retrieve track annotations for regions (e.g., knownGene, cytoband). \\

    \texttt{uniprot} &
    Interface to \textit{UniProt} (UniProtKB/UniRef/UniParc) for protein entries, clusters, and proteomes with flexible search and streaming. &
    \textbf{Purpose:} protein-centric target validation (function, variants, sequences, biological context) with reliable identifiers. \newline
    \textbf{Functions:} UniProtKB search/retrieve; UniRef cluster queries; UniParc archival lookups; gene-centric/proteome-level retrieval; paginated and bulk streaming. \\

    \texttt{txgemma} &
    Therapeutics-specialized predictive toolset (TxGemma) for small molecules and proteins: ADMET, DTI, biologics developability, and interaction tasks. &
    \textbf{Purpose:} accelerate preclinical prioritization by predicting PK/toxicity/bioactivity signals before costly experiments. \newline
    \textbf{Functions:} ADMET profiling (permeability/BBB/P-gp/bioavailability etc.); binding affinity/bioactivity prediction (Kd/Ki/IC\textsubscript{50}/KIBA, receptor panels); antibody developability and antibody--antigen affinity; PPIs/miRNA/immune bindings; synergy predictions (Loewe/Bliss/ZIP/HSA/CSS); reaction yield prediction (e.g., Buchwald--Hartwig). \\

    \texttt{sbdd} &
    Structure-based drug design pipeline integrating protein structure prediction, pocket detection, and generative ligand design/optimization. &
    \textbf{Purpose:} end-to-end in silico hit generation and optimization from target sequence/structure to candidate ligands. \newline
    \textbf{Functions:} protein structure prediction; pocket identification/reranking; de novo generation, substructure inpainting, property optimization (QED/SA); pocket inverse design (e.g., PocketGen). \\

    \texttt{autodock} &
    Docking toolset based on \textit{AutoDock Vina}, covering molecule preparation, grid-box definition, docking execution, and pipeline automation. &
    \textbf{Purpose:} accelerate virtual screening and lead optimization by predicting poses and affinities at scale. \newline
    \textbf{Functions:} ligand/receptor prep (SMILES/PDB to PDBQT); blind/targeted grid generation; run docking; one-shot end-to-end docking pipelines. \\

    \texttt{filesystem} &
    Secure interface for server-side file/directory operations constrained to whitelisted paths (I/O, listing, moving, searching, metadata, editing). &
    \textbf{Purpose:} orchestrate multi-step computational workflows by managing inputs/outputs/logs with strict safety boundaries. \newline
    \textbf{Functions:} read/write files; read multiple files; create/list/move directories; search files; get file info; pattern-based edit with \texttt{dry\_run}; list allowed directories. \\

    \texttt{openbabel} &
    Interface to \textit{Open Babel} for molecular format conversion, structure standardization, filtering, descriptor operations, and advanced chem ops. &
    \textbf{Purpose:} data standardization/prep across heterogeneous molecular formats for downstream docking/QSAR/ML workflows. \newline
    \textbf{Functions:} format conversion (SMILES/SDF/PDB/MOL2, etc.); add/remove H, salts, fragments; coordinate generation; filtering/subsetting (SMARTS, descriptors, dedup); property/metadata ops; conformers, alignment, energy minimization/evaluation. \\

    \texttt{search} &
    General-purpose web search capability via integrated providers (e.g., Tavily and Jina DeepSearch) exposed through MCP. &
    \textbf{Purpose:} automate real-time information gathering (literature, patents, trials, news) for biomedical decision-making. \newline
    \textbf{Functions:} query external search providers; return diverse, programmatically consumable results for downstream synthesis. \\

    \texttt{Hydration Free Energy Predictor} &
    Specialized predictor for hydration free energy ($\Delta G_{\mathrm{hydr}}$) using sPhysNet, taking SMILES or SDF-file inputs. &
    \textbf{Purpose:} fast screening proxy for aqueous solubility (an early ADMET gate) to prioritize candidates. \newline
    \textbf{Functions:} predict $\Delta G_{\mathrm{hydr}}$ from SMILES; predict from SDF path; lightweight high-throughput evaluation. \\

    \texttt{p2rank} &
    Machine learning-based algorithm for ligand binding site prediction and pocket identification on 3D protein structures. &
    \textbf{Purpose:} localize target binding sites to guide downstream structure-based virtual screening and \textit{de novo} generative drug design. \newline
    \textbf{Functions:} pocket detection from PDB inputs; binding site coordinate prediction; pocket re-scoring and ranking based on druggability. \\

\texttt{admet-ai} &
    Machine learning platform for comprehensive Absorption, Distribution, Metabolism, Excretion, and Toxicity (ADMET) profiling of small molecules. &
    \textbf{Purpose:} early-stage \textit{in silico} evaluation of pharmacokinetic properties to de-risk candidate molecules and filter out toxic liabilities. \newline
    \textbf{Functions:} predict diverse ADMET endpoints from SMILES; evaluate physicochemical properties; assess specific toxicity risks (e.g., hERG, AMES, DILI). \\

\texttt{reinvent4} &
    Advanced generative chemistry framework utilizing reinforcement learning for molecular design, structural optimization, and multi-stage filtration. &
    \textbf{Purpose:} automate hit-to-lead expansion and multi-parameter lead optimization while enforcing strict structural and property constraints. \newline
    \textbf{Functions:} hit-to-lead generation (R-group exploration, scaffold hopping); rigorous filtration (structural alerts/REOS, physicochemical properties); iterative lead optimization loops seamlessly coupled with structural docking validation. \\
\end{xltabular}
}

\section{Additional Implementation Details}

\subsection{Domain-Aware Tool Selection in Large Tool Suites}
In the research domain, the number of search tools can be large. Our system, therefore, introduces a two-step internal pattern:
\begin{center}
  \texttt{tool\_search} $\rightarrow$ \texttt{execute\_tool} $\rightarrow$ \texttt{tool\_search} $\rightarrow$ \texttt{execute\_tool} $\rightarrow$ \texttt{final\_answer}.
\end{center}
The alternation constraint reduces failure modes where the model repeatedly ``searches'' without executing, and it encourages explicit, auditable tool choice at each step.

\subsection{Long-Output Summarization Policy}
When tool outputs exceed a length threshold, our system summarizes them before they are reused as context. The summarizer is instructed to surface filepaths/URLs and key numeric values first, so downstream steps can reliably reference evidence without raw long context.

\subsection{Detailed Skill Graph Workflows}
\label{sec:detailed_workflows}

To complement the high-level architecture introduced in Layer B, this section details the internal node execution logic for the four canonical drug discovery stages. These directed acyclic graphs (DAGs) ensure that complex computational pipelines are executed with explicit state management, robust error containment, and strict data contracts.

\begin{figure}[!htp]
\centering
\includegraphics[width=0.95\linewidth]{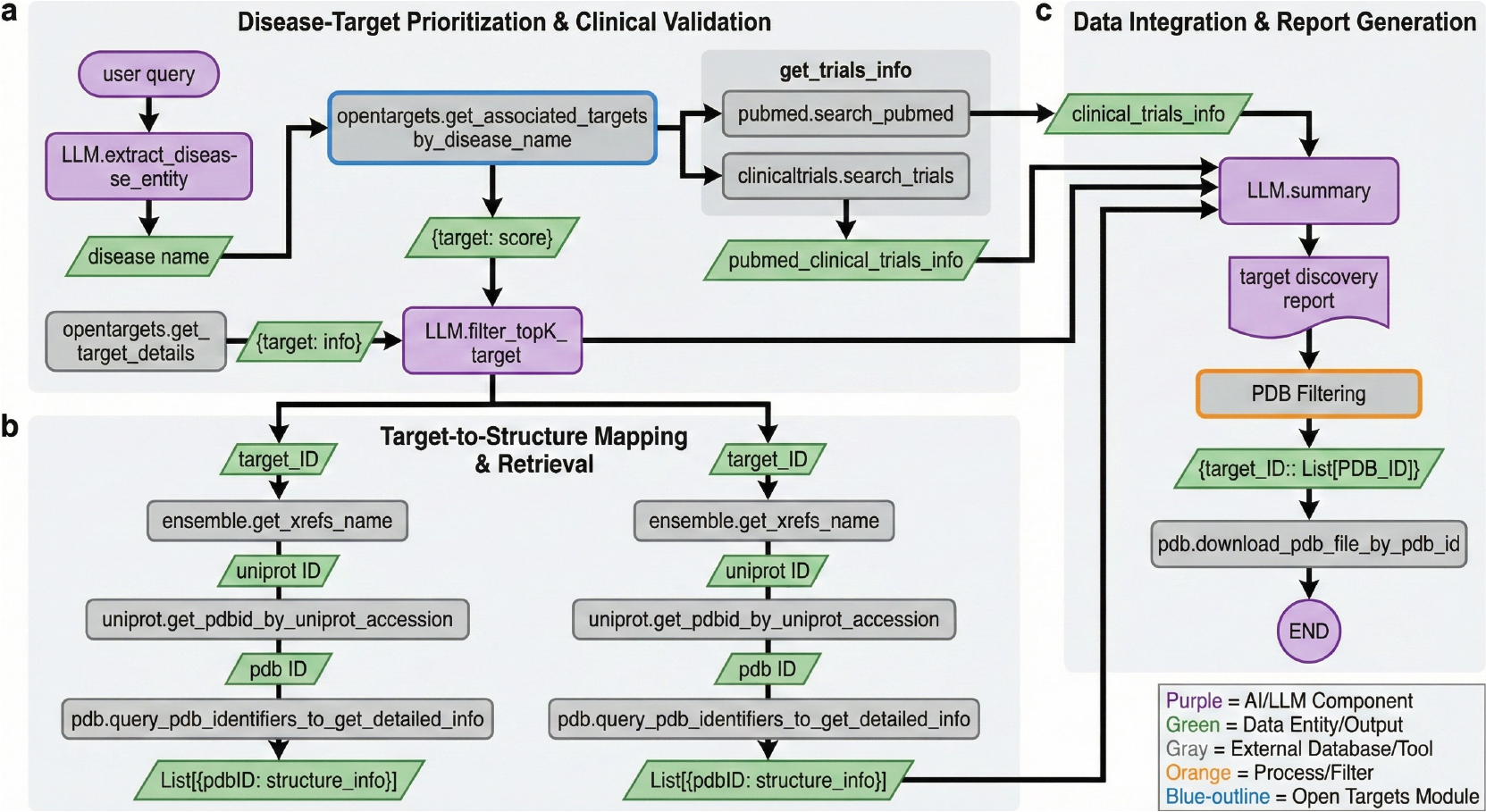}
\caption{Overview of the target identification workflow.}
\label{fig:ti-workflow}
\end{figure}

\paragraph{Target Identification (Figure~\ref{fig:ti-workflow})} 
This workflow translates a high-level natural language disease query into validated protein targets and corresponding structural artifacts. The graph first aggregates multi-source clinical and literature evidence (e.g., Open Targets, PubMed, ClinicalTrials) to prioritize disease-associated targets. Subsequently, it executes a Target-to-Structure Mapping protocol, linking selected targets to standardized identifiers (Ensembl, UniProt) and retrieving high-resolution structural templates from the Protein Data Bank (PDB). The execution concludes with an automated target discovery report and standardized PDB files ready for downstream simulation.

\begin{figure}[!htp]
\centering
\includegraphics[width=0.95\linewidth]{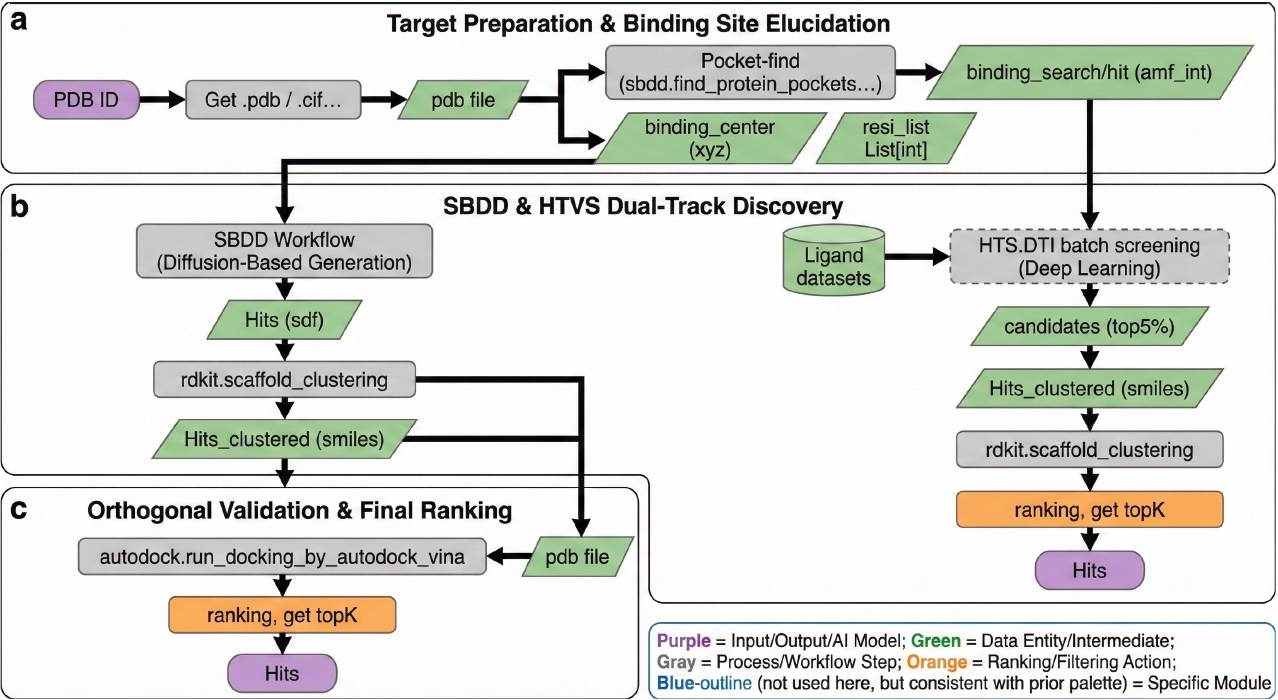}
\caption{Overview of the hit identification workflow.}
\label{fig:hi-workflow}
\end{figure}

\paragraph{Hit Identification (Figure~\ref{fig:hi-workflow})} 
Commencing with the prepared PDB structure, this graph automates binding site elucidation (e.g., pocket finding via P2Rank). It then branches into a parallel dual-track discovery strategy to maximize chemical diversity: (i) a Generative Track utilizing diffusion-based models for \textit{de novo} structure-based drug design (SBDD), and (ii) a Screening Track executing deep learning-based high-throughput virtual screening (HTVS) over massive commercial ligand datasets. Both tracks undergo structural clustering and converge for orthogonal physics-based validation via AutoDock Vina, yielding a ranked list of top-$K$ structural hits.

\begin{figure}[!htp]
\centering
\includegraphics[width=0.55\linewidth]{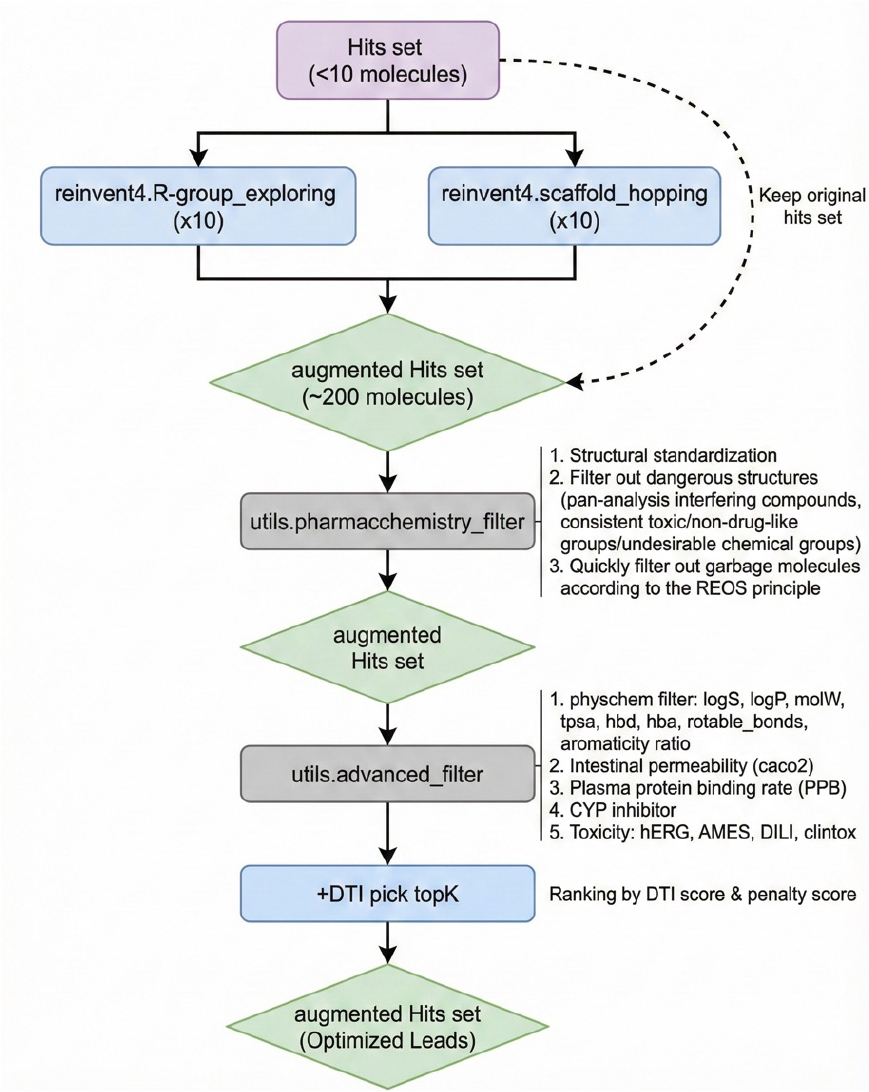}
\caption{Overview of the hit-to-lead workflow.}
\label{fig:h2l-workflow}
\end{figure}

\paragraph{Hit-to-Lead (Figure~\ref{fig:h2l-workflow})} 
This stage aims to transform initial hits into viable lead scaffolds by expanding the localized chemical space. The graph submits the hit set ($<10$ molecules) to a generative chemistry pipeline (e.g., REINVENT4), executing parallel R-group exploration and scaffold hopping. The resulting augmented library ($\sim$200 molecules) is systematically processed through a stringent, multi-tiered filtration cascade. Initial pharmaco-chemistry filters eliminate reactive motifs and pan-assay interference compounds (PAINS, REOS). Surviving candidates then pass through advanced predictive filters evaluating physicochemical properties and critical pharmacokinetic risks (e.g., Caco-2 permeability, plasma protein binding, CYP inhibition, and toxicity). A final Deep Target Interaction (DTI) scoring module outputs the optimized leads.

\begin{figure}[!htp]
\centering
\includegraphics[width=0.95\linewidth]{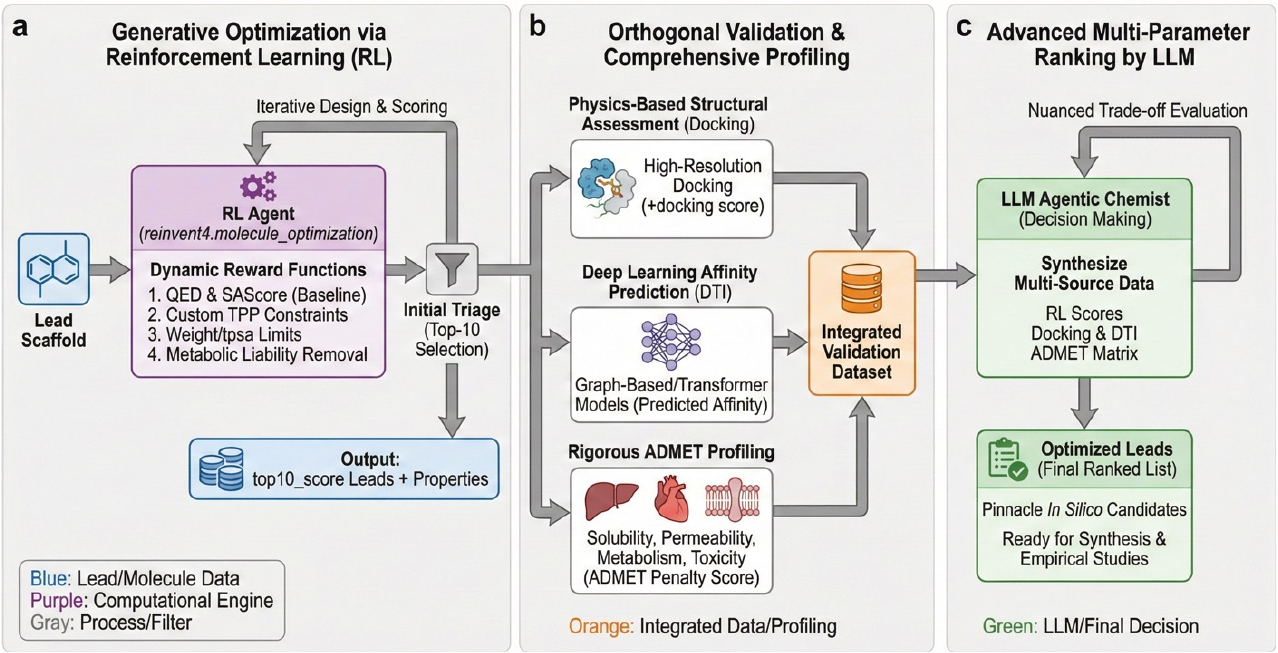}
\caption{Overview of the lead optimization workflow.}
\label{fig:lo-workflow}
\end{figure}

\paragraph{Lead Optimization (Figure~\ref{fig:lo-workflow})} 
The terminal workflow establishes a closed-loop multi-objective reinforcement learning environment to fine-tune lead candidates against strict target product profiles (TPPs). The RL agent iteratively designs and scores molecules based on dynamic reward functions, explicitly balancing properties such as quantitative estimate of drug-likeness (QED), synthetic accessibility (SAScore), and metabolic liabilities. Following an initial triage, candidates undergo rigorous orthogonal profiling, which integrates high-resolution structural docking, deep learning affinity prediction, and comprehensive ADMET penalty scoring. In the final step, an LLM configured as an ``Agentic Chemist'' synthesizes this multi-dimensional validation dataset to perform nuanced trade-off evaluations, producing the final ranked list of \textit{in silico} candidates ready for empirical synthesis.

\section{Cases}
Here we provide comprehensive, step-by-step execution traces of \system{} across three distinct therapeutic areas: Crohn's disease, Parkinson's disease, and sepsis. 
While aggregated metrics evaluate point-wise accuracy, these long-horizon qualitative logs are crucial for illustrating the dynamic behaviors of the governed dual-layer architecture during real-world, multi-day computational pipelines. Specifically, readers should observe the following architectural mechanisms in action:

\begin{itemize}
    \item \textbf{Governed Orchestration \& Intent Routing:} How the Control Plane (Layer A) successfully parses unstructured user intents, translates them into rigid pipeline execution plans, and seamlessly hands off heavy computational tasks to the stateful Workflow Plane (Layer B).
    \item \textbf{Human-in-the-Loop (HITL) Collaboration:} The precise decision boundaries---such as target entity confirmation and PDB structural selection---where the system pauses to synthesize rationales and await expert validation, preventing early-stage hallucinations from propagating downstream.
    \item \textbf{Robust Error Containment:} Explicit instances (e.g., during Sepsis and Crohn's disease hit identification) where localized computational crashes (e.g., AutoDock Vina failures) are caught and handled by the skill graphs, allowing the overarching agentic workflow to proceed without entering infinite loops or catastrophic failure.
    \item \textbf{Multi-Objective Chemical Evolution:} The system's capacity to recognize severe toxicological liabilities (e.g., hERG blockade in the Parkinson's disease trace) during Hit-to-Lead expansion, and subsequently configure a Reinforcement Learning environment to autonomously navigate away from the toxic chemical space.
\end{itemize}

\textit{Disclaimer:} We reiterate that these execution traces represent fully autonomous \textit{in silico} workflows. While the final recommended molecules exhibit highly favorable computational proxy metrics (e.g., binding affinity, QED, ADMET profiles), they remain unverified computational hypotheses. Rigorous \textit{in vitro} and \textit{in vivo} empirical validations are required to substantiate these therapeutic candidates.

\begin{trajectorybox}{Long-horizon Execution Trace: Crohn's Disease Therapeutics}

\textbf{Task Initialization and Intent Routing} \\
The session initiates with a user query requesting drug design for Crohn's disease. The prompt-based intent router in the control plane evaluates the request, recognizes it as a demand for a complete de novo design pipeline without prior structural constraints, and schedules the full end-to-end workflow. The system halts at the first Human-in-the-Loop (HITL) checkpoint, prompting the user to explicitly confirm the disease entity. Upon receiving authorization, the workflow plane activates the Target Identification skill graph.
\medskip

\textbf{Target Identification and Knowledge Retrieval} \\
The research agent queries the Open Targets platform via the standardized Model Context Protocol, retrieving 25 potential protein targets associated with Crohn's disease. The agent filters and submits the top 20 candidates to the large language model for qualitative evaluation. The model successfully prioritizes a top-5 list, including targets such as NOD2 (a primary genetic risk factor) and ITGA4 (integrin alpha-4, clinically validated by vedolizumab). The agent then queries the Protein Data Bank (PDB) to fetch structural data for these targets. It identifies the PDB structure 3V4V for ITGA4, which represents the complete $\alpha_4\beta_7$ integrin heterodimer complexed with a therapeutic antibody fragment. The system pauses at an HITL gate to present the target rationale and structural options. The user selects the 3V4V structure to advance the pipeline.
\medskip

\textbf{Hit Identification via Generative Design and Error Containment} \\
To localize the binding site, the computation worker invokes the P2Rank algorithmic tool. The tool scans the 3V4V structure and identifies 42 potential binding pockets, ultimately selecting the top-ranked pocket centered at coordinates [2.3, 58.4, 5.1] with high confidence. Unlike the virtual screening approach, the system deploys the DiffSBDD generative diffusion model to perform de novo ligand design within this specific pocket geometry, generating 49 novel molecular structures. To ensure chemical diversity, the system applies structural clustering to narrow the pool down to 20 representative hits. The computation worker then evaluates these candidates using AutoDock Vina. Crucially, the execution log reveals that several generated structures cause the docking engine to fail. The stateful skill graph catches these execution exceptions locally, prevents the errors from propagating to the supervisor agent, and allows the pipeline to continue. The top successfully docked molecule (SMILES: Cc1cc2nc(N[C@H]3CCNCC3=O)cnc2cc1C(=O)O) achieves a strong binding score of -9.0 kcal/mol.
\medskip

\textbf{Hit-to-Lead Expansion and Pharmacochemistry Filtering} \\
The user confirms the 8 most promising hits at the subsequent HITL checkpoint. The workflow plane submits these molecules to the REINVENT4 framework for chemical space expansion using parallel R-group exploration and scaffold hopping techniques. This process yields a broad library of candidates that are subsequently passed through a rigorous multi-stage filtering pipeline. The system programmatically applies structural alerts (PAINS, REOS) and executes quantitative predictive models (ADMET-AI) to evaluate Caco-2 cell permeability, plasma protein binding, and critical toxicity risks, including hERG blockade, AMES mutagenicity, and drug-induced liver injury (DILI). The rigorous filtration successfully distills the expanded library down to 5 pristine lead candidates, all of which achieve a penalty score of zero, indicating an optimal balance of physicochemical properties and safety.
\medskip

\textbf{Lead Optimization and Final Candidate Selection} \\
In the final stage, the lead optimization agent configures a reinforcement learning environment to refine the 5 validated leads. The reward function is explicitly designed to maximize the quantitative estimate of drug-likeness (QED) and the synthetic accessibility score (SAScore) while maintaining strong binding affinity within the defined 20 $\times$ 20 $\times$ 20 \AA\ docking box. The system generates thousands of iterations and selects the top 10 candidates for final docking validation. The workflow correctly penalizes molecules that exhibit good physicochemical properties but fail the final structural docking. The terminal candidate (SMILES: CC1(C(=O)O)CC(c2cccc(Oc3cnccn3)c2)C1) emerges as the top recommendation, achieving a highly favorable docking score of -8.8 kcal/mol, an excellent QED value of 0.933, and a perfect synthetic accessibility score of 1.0, making it a highly viable candidate for experimental synthesis.

\end{trajectorybox}
\vskip 20pt

\begin{trajectorybox}{Long-horizon Execution Trace: Parkinson's Disease Therapeutics}

\textbf{Task Initialization and Intent Routing} \\
The session commences with the user query requesting the design of novel therapeutics for Parkinson's disease. The prompt-based intent router analyzes the unstructured request, classifies it as an end-to-end drug design task, and schedules the full sequential workflow. The system encounters its first Human-in-the-Loop (HITL) checkpoint, pausing execution to explicitly verify the disease entity before launching any computational tools. Upon user confirmation, the trajectory proceeds to the first stage.
\medskip

\textbf{Target Identification and Structural Mapping} \\
The research worker queries the OpenTargets database via the standardized MCP bus, retrieving 25 candidate proteins associated with Parkinson's disease. An integrated large language model evaluates the top 20 candidates based on genetic association, biological relevance, and clinical tractability. The system successfully prioritizes the leucine-rich repeat kinase 2 (LRRK2, ENSG00000188906) as the primary target, citing its strong genetic linkage, high druggability as a kinase, and critical role in vesicle trafficking. Subsequently, the system maps LRRK2 to the Protein Data Bank (PDB), evaluating multiple structural templates. It selects the 8TXZ structure, a 3.05 \AA\ cryo-EM model capturing the kinase domain bound to an MLi-2 inhibitor. The system halts at an HITL gate, presenting the rationale and structural data. The user approves the 8TXZ selection to advance the pipeline.
\medskip

\textbf{Hit Identification via High-Throughput Virtual Screening} \\
To locate the binding pocket, the system presents two algorithmic pathways to the user: a deep-learning-based geometric scan (DiffSBDD) or a legacy reference-ligand method. The user selects the legacy mode. The system programmatically extracts the coordinates of the co-crystallized A1N ligand, establishing the binding center at [161.2, 205.8, 151.3] with a bounding box of 17.5 $\times$ 24.7 $\times$ 28.0 \AA. The computation worker then executes a massive high-throughput virtual screening over a library of 377,760 molecules. Utilizing the LigUnity drug-target interaction model, the system scores the binding affinity of the candidates. The top 5\% of the pool undergoes structural clustering to eliminate redundancy. The system outputs a diverse list of 10 hit compounds, where the top candidate (SMILES: CC1CN(c2cc(-c3n[nH]c4ccc(OC5(C)CC5)cc34)ncn2)CC(C)O1) achieves an exceptional prediction score of 0.972. The user validates the hit list and authorizes the transition to the next phase.
\medskip

\textbf{Hit-to-Lead Expansion and Liability Detection} \\
The workflow plane initiates the hit-to-lead stage by submitting the 10 initial hits to a generative chemistry pipeline powered by REINVENT4. Parallel workflows execute R-group exploration and scaffold hopping, successfully expanding the chemical space to 844 unique candidates. The system then routes these molecules through rigorous pharmacochemistry filters. It programmatically eliminates reactive pan-assay interference compounds (PAINS) and applies the rapid elimination of swill (REOS) criteria. The surviving candidates are subjected to ADMET evaluation. The execution log reveals a critical systemic insight: while the top 5 expanded leads maintain high interaction scores (up to 0.958), the system actively flags severe toxicological liabilities across these scaffolds, explicitly warning of hERG channel blockade, drug-induced liver injury (DILI) risks, and multi-isoform CYP inhibition. The system assigns heavy penalty scores to these molecules and presents the risk report to the user, who instructs the system to aggressively optimize out these liabilities in the final stage.
\medskip

\textbf{Lead Optimization and Multi-Parameter Convergence} \\
Recognizing the toxicity flaws, the lead optimization agent configures a multi-parameter reinforcement learning environment. The reward function is heavily constrained by five critical pharmacokinetic objectives: solubility, non-toxicity, blood-brain barrier (BBB) permeability, hERG safety, and bioavailability. Over thousands of generative iterations, the system iteratively navigates away from the toxic chemical space to discover a completely novel scaffold. The final optimized candidate (SMILES: CC1=NC=C(C=N1)C2=NNC3=C2C=C(C=C3)CC4(CC4)OC) demonstrates a remarkable structural evolution. It achieves an AutoDock Vina binding energy of -8.924 kcal/mol while simultaneously securing a top-tier hERG safety score of 0.399 and a BBB permeability score of 0.748. The final system report explicitly benchmarks this novel molecule against the established clinical compound DNL-201. The agent's analysis indicates that the generated candidate exhibits a highly favorable multi-parameter in silico profile that meets or exceeds the computational proxy metrics of the clinical benchmark across the evaluated ADMET parameters. However, we strictly interpret these results as surrogate-driven hypotheses; rigorous in vitro and in vivo validations, alongside rigorous uncertainty quantification of the underlying ML models, are necessary to substantiate these computational findings.
\end{trajectorybox}
\vskip20pt

\begin{trajectorybox}{Long-horizon Execution Trace: Sepsis Therapeutics}

\textbf{Task Initialization and Intent Routing} \\
The session begins with a generalized user query requesting drug discovery for sepsis, a severe systemic inflammatory response. The prompt-based intent router analyzes the request, infers the need for a complete multi-stage pipeline, and configures the end-to-end execution mode. The system halts at the initial Human-in-the-Loop (HITL) checkpoint to request explicit confirmation of "Sepsis" as the target disease. Following user validation, the target identification skill graph is instantiated.
\medskip

\textbf{Target Identification and Structural Selection} \\
The research worker queries the integrated Open Targets and UniProt databases, retrieving 25 potential protein targets associated with sepsis pathophysiology. The large language model evaluates the evidence and isolates a high-priority top-5 list, including targets such as TLR4 (central to lipopolysaccharide recognition) and PTGS2 (COX-2, critical for inflammatory prostaglandin production). However, the system designates the $\beta_2$-adrenergic receptor (ADRB2, ENSG00000169252) as the top-ranked candidate. The generated rationale highlights the direct biological relevance of ADRB2 in modulating inflammatory responses and vascular tone during septic shock, alongside its high clinical tractability. The system subsequently maps ADRB2 to available structures and selects PDB 7BZ2, a 3.82 \AA\ cryo-EM structure capturing the active signaling state of the receptor bound to formoterol and a Gs protein. The user approves the target and structural selection at the corresponding HITL gate.
\medskip

\textbf{Hit Identification via De Novo Generation} \\
With the 7BZ2 structure validated, the computation worker executes the hit identification graph. The P2Rank algorithm scans the protein surface, identifying 19 potential binding sites. It selects the highest-scoring site (Pocket1) centered at [104.3, 106.1, 71.6] with an exceptional confidence probability of 0.992. The system then invokes the DiffSBDD generative model to design 48 novel molecules specific to this pocket's geometry. To maintain a computationally efficient yet diverse search space, the system clusters the generated scaffolds and isolates 20 representative molecules for docking. AutoDock Vina evaluates the candidates, and the system gracefully handles and logs 7 molecular docking failures without crashing the overarching workflow. The top successful hit (SMILES: CC[C@H]1COC[C@H]1c1ccc(OCC2=NC[C@H]3CCON3C2=O)cc1) achieves an excellent predicted binding affinity of -8.7 kcal/mol.
\medskip

\textbf{Hit-to-Lead Expansion and Zero-Penalty Filtration} \\
Following user confirmation of the top 13 initial hits, the workflow progresses to the hit-to-lead stage. The system deploys the REINVENT4 framework, executing parallel R-group exploration and scaffold hopping to generate a massive library of analog structures. The computation worker then subjects this expanded chemical space to a stringent, multi-tiered filtering cascade. The process systematically strips out compounds containing pan-assay interference (PAINS) motifs and applies the rapid elimination of swill (REOS) rules. The surviving structures undergo comprehensive ADMET screening using integrated deep-learning predictors. Crucially, the system successfully filters out molecules exhibiting hERG blockade liability, AMES mutagenicity, and drug-induced liver injury (DILI) risks. This exhaustive distillation yields five optimal lead candidates, all of which achieve a perfect penalty score of zero, indicating an excellent balance of physicochemical properties and safety.
\medskip

\textbf{Lead Optimization and Robust Ranking} \\
In the final stage, the lead optimization agent initializes a reinforcement learning environment to fine-tune the five validated leads. The objective function is programmed to explicitly optimize the quantitative estimate of drug-likeness (QED) and the synthetic accessibility score (SAScore). Following extensive generative iteration, the system selects the top 10 candidates based on property scores and submits them for final validation, docking against the 7BZ2 pocket. The execution trace reveals that two candidates experience docking failures, while two others yield anomalous zero scores. The stateful ranking logic successfully accounts for these physical validation failures, down-ranking the problematic structures despite their high QED values. The final recommended molecule (SMILES: Cc1ccc(Cc2ccc(S(N)(=O)=O)cc2C)c(C)c1) emerges as the clear optimal choice, achieving the strongest binding affinity (-8.4 kcal/mol), an exceptional QED of 0.944, and a perfect synthetic accessibility score of 1.0.

\end{trajectorybox}